%% file: main.tex
\documentclass[runningheads]{llncs}

\usepackage[mobile]{eccv}

\usepackage{eccvabbrv}

\usepackage{graphicx}
\usepackage{pifont}
\newcommand{\cmark}{\ding{51}}
\newcommand{\xmark}{\ding{55}}
\usepackage{booktabs}
\usepackage[table]{xcolor}
\usepackage[table]{xcolor}
\usepackage{multirow}
\usepackage{caption}
\usepackage{wrapfig}

\usepackage[accsupp]{axessibility}  

\usepackage{hyperref}

\usepackage{orcidlink}

\begin{document}

\title{Head Forcing: Long Autoregressive Video Generation via Head Heterogeneity}

\titlerunning{Head Forcing}

\author{Jiahao Tian\inst{1} \and
Yiwei Wang\inst{2} \and
Gang Yu\inst{3} \and Chi Zhang\inst{1}\textsuperscript{$\dagger$}}

\authorrunning{J. Tian et al.}

\institute{AGI Lab, Westlake University \and University of California at Merced  \and StepFun
}

\maketitle
{
  \renewcommand{\thefootnote}{\fnsymbol{footnote}}
  \setcounter{footnote}{0}
  \footnotetext{$\dagger$ denotes corresponding author.}
}

\input{sections/abstract}
\input{sections/introduction}
\input{sections/related_work}

\input{sections/method}
\input{sections/experiments}
\input{sections/conclusion}
%
%
\bibliographystyle{splncs04}
\bibliography{main}

\clearpage
\renewcommand\thefigure{S\arabic{figure}}
\setcounter{figure}{0}
\renewcommand\thetable{S\arabic{table}}
\setcounter{table}{0}
\renewcommand\theequation{S\arabic{equation}}
\setcounter{equation}{0}
\pagenumbering{arabic}
\renewcommand*{\thepage}{S\arabic{page}}
\setcounter{footnote}{0}
\setcounter{page}{1}
\appendix

\begin{center}
{\LARGE \textbf{Supplementary Material}}
\end{center}
\input{sections/supp}


\end{document}

%% file: sections/abstract.tex
\begin{abstract}
Autoregressive video diffusion models support real-time synthesis but suffer from error accumulation and context loss over long horizons. We discover that attention heads in AR video diffusion transformers serve functionally distinct roles as local heads for detail refinement, anchor heads for structural stabilization, and memory heads for long-range context aggregation, yet existing methods treat them uniformly, leading to suboptimal KV cache allocation. We propose \textbf{Head Forcing}, a training-free framework that assigns each head type a tailored KV cache strategy: local and anchor heads retain only essential tokens, while memory heads employ a hierarchical memory system with dynamic episodic updates for long-range consistency. A head-wise RoPE re-encoding scheme further ensures positional encodings remain within the pretrained range. Without additional training, Head Forcing extends generation from 5 seconds to minute-level duration, supports multi-prompt interactive synthesis, and consistently outperforms existing baselines. Project Page: \url{https://jiahaotian-sjtu.github.io/headforcing.github.io/}.
  \keywords{Autoregressive video generation \and Long video generation \and KV cache}
\end{abstract}

%% file: sections/introduction.tex
\section{Introduction}
\label{sec:intro}

Autoregressive (AR) video diffusion models~\cite{huang2025self, yin2025slow, teng2025magi, chen2024diffusion, chen2025skyreels} have recently emerged as a promising paradigm for long video generation. Unlike bidirectional video diffusion models~\cite{wan2025,yang2024cogvideox,villegas2022phenaki,polyak2024movie,ho2022video,ho2022imagen,singer2022make,kong2024hunyuanvideo,hacohen2024ltx,gupta2024photorealistic,blattmann2023align,blattmann2023stable,chen2023videocrafter1,girdhar2023emu,ma2024latte,zhang2025show,henschel2024streamingt2v,zhang2023i2vgen} that process all frames simultaneously and suffer from quadratic attention cost over temporal length, AR models generate video in a sequential fashion with KV caching, offering inherent scalability to extended durations and enabling real-time streaming synthesis. These advantages have positioned AR video diffusion as the dominant framework for pushing video generation beyond the short-clip regime.

Despite their promise, AR video diffusion models face two challenges that grow with generation length. First, error accumulation produces color drift and detail loss. Second, bounded KV caches discard earlier context, leading to identity drift and scene forgetting.
Training-based extensions such as Rolling Forcing~\cite{liu2025rolling}, LongLive~\cite{yang2025longlive}, and Self-Forcing++~\cite{cui2025self} achieve minute-level consistency by anchoring generation to initial frames with periodic DMD retraining, but can suffer from static artifacts due to first-frame bias. Other approaches~\cite{ji2025memflow,yu2025context,sun2025worldplay} leverage memory mechanisms for long-range consistency but require complex architectural modifications. All such methods incur considerable training cost.
More recently, training-free strategies~\cite{yesiltepe2025infinity, yi2025deep, yang2026stableworld, li2026rolling} have emerged, typically leveraging attention sinks with RoPE adjustments for stable extended synthesis. However, like training-based methods, they rely solely on attention sinks from the first few frames while discarding intermediate context. This prevents tracking subjects and scenes emerging mid-sequence, yielding homogeneous or static outputs under single-prompt video generation and inconsistent transitions under multi-prompt video generation settings. Moreover, uniform treatment of all attention heads overlooks their distinct functional roles whose
impact has not been systematically examined.

To this end, we first profile the multi-head self-attention layers of AR video DiTs, revealing three functionally distinct head types in long generation. (1) \textbf{Memory heads} attend broadly across the full context to capture narrative elements and sustain long-term memory. (2) \textbf{Local heads} focus on the current block and its immediate neighborhood for detail refinement and motion continuity. (3) \textbf{Anchor heads} exhibit significantly elevated first-frame attention, using the initial frame as an anchor to prevent visual collapse.
Uniform KV cache allocation~\cite{yi2025deep, yesiltepe2025infinity} thus creates a fundamental mismatch: heads rarely utilizing historical context receive excessive budget, wasting computation and introducing noise, while memory heads lack sufficiently rich historical evidence. This mismatch intensifies during long rolling generation, increasing drift accumulation and consistency degradation.

Based on this insight, we propose a training-free, head-wise framework built on AR video DiTs for long, controllable video generation that allocates KV cache according to each head's functional role. \textbf{Local heads} access only the current block and nearest neighboring frames for detail synthesis and motion continuity. \textbf{Anchor heads} access to the first few latent frames alongside the same local tokens to stabilize generation. Both designs conserve cache budget for memory heads while suppressing error propagation. For \textbf{memory heads}, we replace conventional sliding windows with attention sinks by a hierarchical memory system with dynamic updates: (1) fast memory for the immediate neighbouring scenery, and (2) episodic memory storing representative, clean frames from distinct scenes. Episodic memory updates dynamically based on scene novelty and undergoes prompt-guided compression upon reaching capacity. Finally, we propose head-wise RoPE re-encoding tailored to this cache allocation, ensuring frame-level relative positional encoding within each head remain consistently within the pretrained range to reduce positional-inconsistency errors.

Built upon pretrained AR video DiTs, e.g., Self Forcing~\cite{huang2025self},  our method extends video generation from 5 seconds to minute-level duration without noticeable quality degradation and further enables long horizon, contextually consistent, prompt-guided interactive video synthesis.
In summary, our contributions are:
\begin{itemize}
    \item We profile attention heads in AR video DiTs and identify three functional head types: memory heads, local heads and anchor heads, offering a new perspective on temporal modeling in AR video generation.

    \item We propose a training-free, head-wise framework that assigns heterogeneous KV cache strategies by head role: local heads access only the current block and nearest neighbors for detail synthesis and motion continuity, anchor heads preserve the first few latent frames to stabilize generation, and memory heads adopt a hierarchical fast and episodic memory with dynamic updates to preserve long-range consistency with reduced error accumulation.
    
    \item We further introduce a head-wise RoPE re-encoding strategy that keeps frame-level relative positions within the pretrained range, mitigating positional encoding drift.
\end{itemize}

%% file: sections/related_work.tex
\section{Related Work}
\label{sec:related_work}
\noindent\textbf{Video Diffusion Models and Autoregressive Video Diffusion Models.} Following image diffusion models~\cite{ho2020denoising,rombach2022high,song2020denoising,liu2022flow,lipman2022flow}, video diffusion models~\cite{wan2025,yang2024cogvideox,villegas2022phenaki,polyak2024movie,ho2022video,ho2022imagen,singer2022make,kong2024hunyuanvideo,hacohen2024ltx,gupta2024photorealistic,blattmann2023align,blattmann2023stable,chen2023videocrafter1,girdhar2023emu,ma2024latte,zhang2025show,henschel2024streamingt2v,zhang2023i2vgen} typically process all frames simultaneously, employing bidirectional attention throughout the denoising process. With large-scale training, these models have achieved high-resolution video generation that maintains coherent visual appearance and motion continuity over dozens of frames. However, they remain fundamentally limited to short durations because they are trained exclusively on short clips. Extending these architectures to long videos is challenging due to prohibitively long token sequences.
Recently, a rapidly growing number of AR video diffusion models~\cite{weng2024art,hu2024acdit,gu2025long,chen2024diffusion,gao2024ca2,guo2025long,song2025history,po2025bagger,liu2025rolling,yi2025deep,lu2025reward,cui2025self,guo2025end,ji2025memflow,yu2025videossm,zhang2025frame,bruce2024genie,ren2025next,wang2024loong,weissenborn2020scalingautoregressivevideomodels,yuan2025lumos,yesiltepe2025infinity,xiao2025knot,li2025joyavatar,zhou2026videomemory} combine diffusion modeling with autoregressive prediction to support long-horizon or streaming video generation.
MAGI-1~\cite{teng2025magi} and Diffusion Forcing~\cite{chen2024diffusion} generate videos
chunk-by-chunk with progressive denoising, enabling streaming generation. CausVid~\cite{yin2025slow} converts a pretrained bidirectional diffusion transformer into an AR generator with KV caching. Building on
these ideas, Self Forcing~\cite{huang2025self} addresses the train-inference
mismatch by conditioning the model on its own generated
frames.

\noindent\textbf{Long Video Generation.}
Bidirectional video diffusion models~\cite{wan2025, yang2024cogvideox,villegas2022phenaki,polyak2024movie,ho2022video,ho2022imagen,kong2024hunyuanvideo,hacohen2024ltx,blattmann2023stable,chen2023videocrafter1,girdhar2023emu,ma2024latte,tian2025freeloc} achieve impressive short video quality but struggle with long videos due to the quadratic attention cost, constraining scalability to extended temporal horizons.
AR video diffusion models such as CausVid~\cite{yin2025slow} and Self Forcing~\cite{huang2025self} offer a more scalable paradigm by generating frames sequentially, yet suffer from long-horizon drift caused by error accumulation. To mitigate this, LongLive~\cite{yang2025longlive} and Rolling Forcing~\cite{liu2025rolling} introduce attention sinks with sliding-window KV caches; Self-Forcing++~\cite{cui2025self} extends the DMD~\cite{yin2024one,yin2024improved} formulation through long streaming tuning; Other approaches~\cite{xiao2025worldmem, ji2025memflow,yu2025videossm,chen2026context,zhu2025memorize,zhang2025packing,yu2025context} leverage memory mechanisms to enhance long-range consistency. Despite their effectiveness, these strategies typically require complex architectural modifications and extensive re-training, limiting practical applicability.
More recently, training-free methods have emerged to extend AR video generation length. Deep Forcing~\cite{yi2025deep} and Rolling Sink~\cite{li2026rolling} employ attention sink based cache stabilization for long rollout, while Infinity-RoPE~\cite{yesiltepe2025infinity} and LoL~\cite{cui2026lol} modify RoPE at inference time for temporal encoding stability. StableWorld~\cite{yang2026stableworld} applies dynamic frame eviction to reduce accumulated errors in interactive generation.
However, these approaches apply uniform KV cache strategies across all attention heads, disregarding their functionally distinct roles. This yields suboptimal allocation that introduces noise while discarding important context, and the absence of long-term memory mechanisms causes consistency degradation over extended horizons. Our head-wise framework with hierarchical memory addresses both limitations.

%% file: sections/method.tex
\section{Method}
\label{sec:method}
\begin{figure}[t!]
\vspace{-15pt}
    \centering
    \includegraphics[width=1\linewidth]{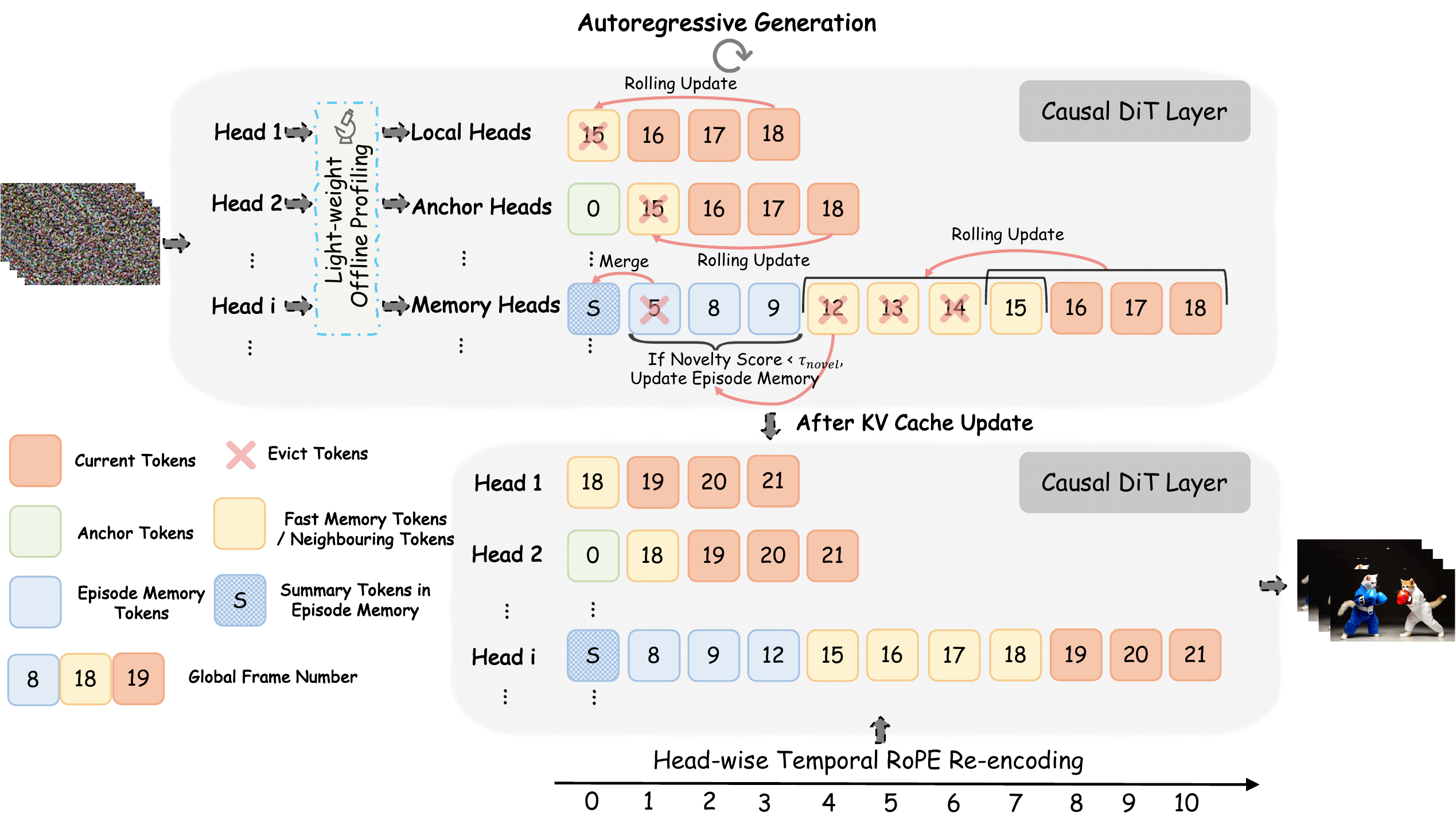}
    \caption{\textbf{Overview of Head Forcing. } Attention heads are profiled offline into local, anchor, and memory heads, each receiving a tailored KV cache strategy. Memory heads are equipped with a hierarchical memory system with dynamic updates. Head-wise RoPE re-encoding ensures positional consistency across all heads.}
    \label{fig:pipeline}
   \vspace{-15pt} 
\end{figure}

We first profile the functional role of each attention head during long-rollout AR video diffusion. Based on this analysis, as shown on Fig.~\ref{fig:pipeline}, we propose \textbf{Head Forcing}, a training-free, head-wise framework with RoPE re-encoding built on pretrained AR video DiTs (e.g., Self Forcing~\cite{huang2025self}), enabling robust minute-level video generation and long-term consistent prompt-guided interactive generation.

\subsection{Preliminary}
\noindent\textbf{Autoregressive Video Diffusion Model.}
In AR video diffusion models, the joint distribution over an $N$-frame video $x_{1:N}$ is factorized as
\begin{equation}
    p(x^{1:N}) = \prod_{i=1}^{N} p\!\left(x^i \mid x^{<i}\right).
    \small
\end{equation}
A key challenge in AR video diffusion is the training-inference discrepancy, which Self Forcing~\cite{huang2025self} mitigates through a self-rollout mechanism that aligns training with inference via diffusion-based conditional formulation. In block-wise AR video generation, the sequence of $N$ frames is decomposed into $N/f$ successive blocks of $f$ latent frames each. Each block is denoised from Gaussian noise using a few-step diffusion model conditioned on prior frames through sliding-window attention with fixed cache size $W$ over timesteps $\{0,1,\ldots,T\}$. A KV cache of fixed capacity stores recent-frame representations under a FIFO eviction policy.
Under this scheme, at timestep $t$, for the $i$-th block (frames $fi{-}f{+}1,\ldots,fi$), the self-attention layers operate as:
\begin{equation}
    \mathrm{SelfAttention}=\mathrm{softmax}\!\left(Q_i \left[K_{\mathrm{cache}}, K_i \right]^{\top}/\sqrt{d} \right)\left[ V_{\mathrm{cache}}, V_i \right],
    \label{eq:mha_concat}
    \small
\end{equation}
where $Q_i$, $K_i$, $V_i$ are the query, key, and value of the current block, $K_{\mathrm{cache}}$ and $V_{\mathrm{cache}}$ are the cached keys and values, $[\cdot]$ denotes sequence-wise concatenation, and $d$ is the key dimensionality.

\begin{wrapfigure}{r}{0.75\textwidth}
    \vspace{-.85cm}
    \centering
    \includegraphics[width=\linewidth]{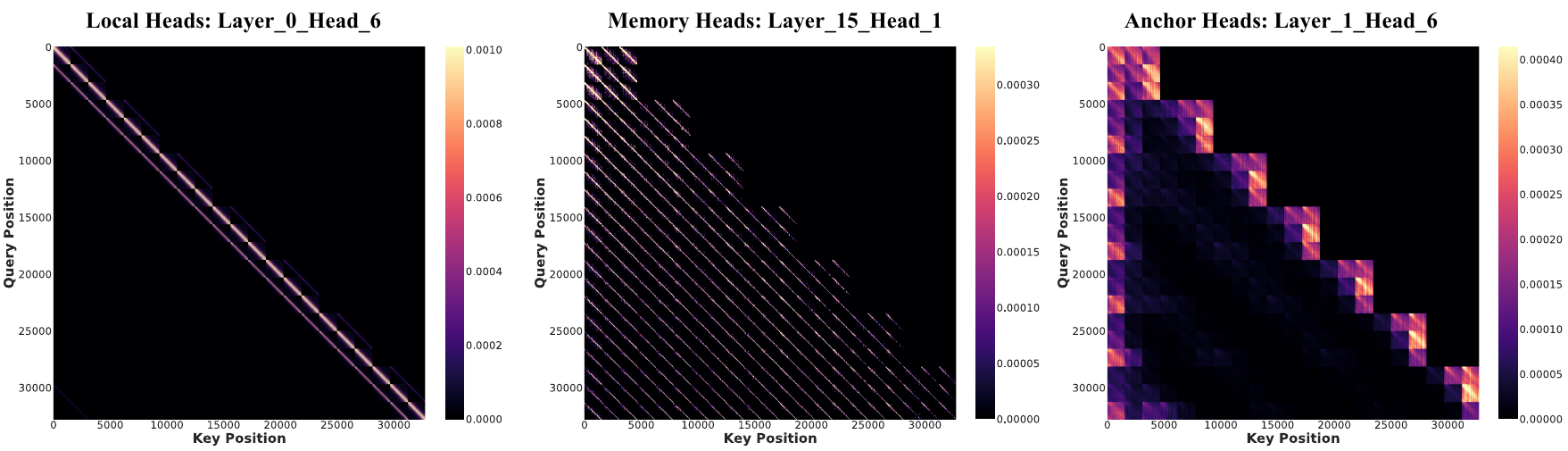}
    \caption{
    \textbf{Representative attention patterns for different attention heads.}
    }
    \label{fig:attn_map}
    \vspace{-2.2em}
\end{wrapfigure}

\subsection{Profile the Different Role of Attention Heads}
We examine the AR video DiT generation process to identify properties exploitable for long video generation. Specifically, we profile how the $i$-th block interacts with preceding frames across attention heads, characterizing their context dependency patterns and roles in long generation.

\noindent\textbf{Long Video Profiling Process.}
We profile Self Forcing~\cite{huang2025self} in the main paper and report results on other AR video models in the \emph{supplementary}.
To characterize how each attention heads attend to the historical KV cache in long video generation, we extend the AR rollout beyond the training horizon using a sliding window over the KV cache together with a sink frame. This allows us to profile each head's role for long video generation.
Following Self Forcing~\cite{huang2025self}, each block contains $f{=}3$ frames. For the $i$-th block, let $Q_i$, $K_i$, and $V_i$ denote the query, key, and value tokens of the current block. The per-head attention map is
$A=\mathrm{softmax}\!\left(Q_i [K_{\mathrm{pre}},\,K_i]^{\top}/{\sqrt{d}}\right)$,
where $K_{\mathrm{pre}}$ denotes cached key tokens from all previously generated frames and $[\ ,\ ]$ denotes token-dimension concatenation. Thus $A\in\mathbb{R}^{3s\times 3is}$, where $s$ is the token count per frame and $3i$ the total frame count including previous and current blocks. Representative per-head attention maps within one sliding window are shown in Fig.~\ref{fig:attn_map}.

\begin{wrapfigure}{r}{0.7\textwidth}
    \vspace{-.85cm}
    \centering
    \includegraphics[width=\linewidth]{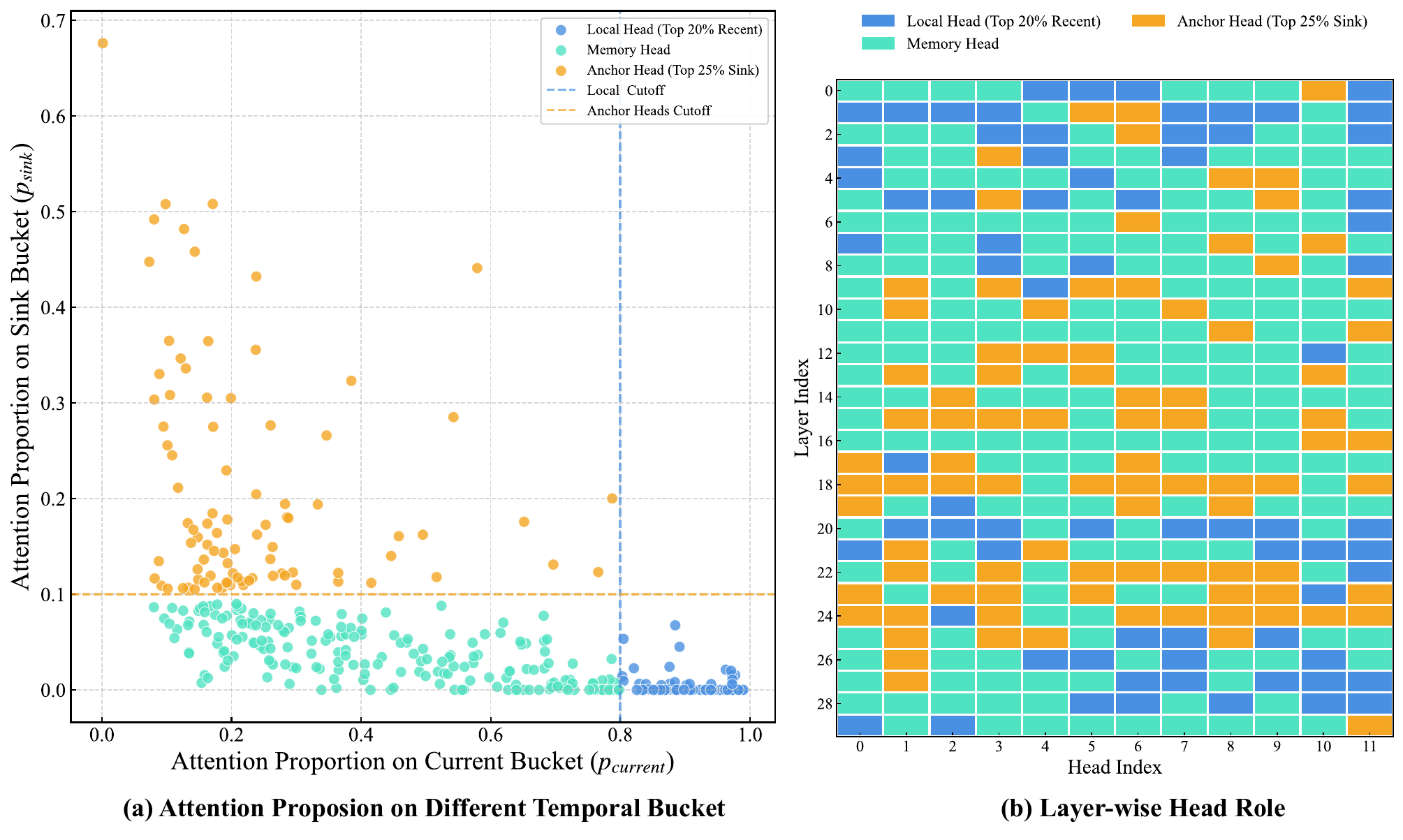}
    \caption{
    \textbf{Attention head profiling}. (a) Attention proportion for each head, showing clear clustering into local, anchor, and memory heads. (b) Layer-wise head role distribution across all transformer layers.}
    \label{fig:headrole}
    \vspace{-2.2em}
\end{wrapfigure}

Beyond visual inspection, we further quantify how the current block allocates attention to different temporal regions of the KV cache. 
Specifically, we partition the full attention context into three temporal buckets: \emph{sink bucket} (first frame, $[0,\,s)$), \emph{middle bucket} (intermediate historical frames, $[s,\,(3i-3)s)$), and \emph{current bucket} (current block, $[(3i-3)s,\,3is)$). We compute the attention proportion for each bucket by aggregating attention weights over all its token indices:
\begin{equation}
    p_b=\frac{1}{3s}\sum_{m=1}^{3s}\sum_{n\in I_b}A_{mn},
    \label{eq:frame_group_score}
    \small
\end{equation}
where $I_b$ denotes the token index set of bucket $b$, and $b\in\{\mathrm{sink},\mathrm{middle},\mathrm{current}\}$. All profiling statistics in Fig.~\ref{fig:headrole} are averaged over multiple prompts, AR steps, and denoising steps. 
We then group attention heads by their attention patterns into \textbf{local heads}, \textbf{anchor heads}, and \textbf{memory heads}, and provide empirical evidence for their functional roles.

\begin{figure}[htbp]
\vspace{-8pt}
  \centering
  \begin{minipage}{0.6\textwidth}
    \centering
    \includegraphics[width=\linewidth]{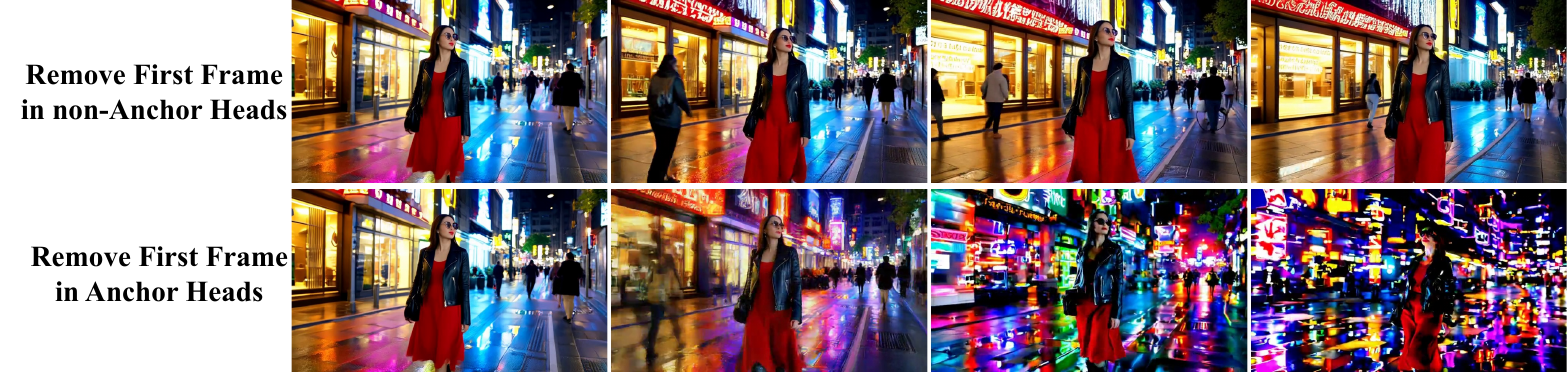}
    \captionof{figure}{Effect of removing the first latent frame from anchor heads vs.\ non-anchor heads. }
     \label{fig:anchor}
  \end{minipage}
  \hfill
\begin{minipage}{0.38\textwidth}
  \centering

  \resizebox{\linewidth}{!}{%
    \begin{tabular}{lccc}
        \toprule
        Setup& Quality&  Consistency& Total\\
        \midrule
        Self Forcing & 84.82 & 96.78 & 83.92 \\
        Pruning in non-Local Heads& 81.58 & 95.82 & 80.74\\
        Pruning in Local Heads& 84.78 & 96.80 & 83.88 \\
        \bottomrule
    \end{tabular}
  }
  \captionof{table}{Ablation on local head KV cache pruning.}
    \label{tab:local}
\end{minipage}
  \vspace{-15pt}
\end{figure}

\noindent\textbf{\emph{Local Heads.}}
As shown in Fig.~\ref{fig:attn_map}, these attention heads place overwhelmingly high attention on the current block and its immediate neighboring frame, even under long historical context, suggesting that they mainly support local detail generation and short-range temporal continuity. To identify them, we measure for each head the proportion of attention assigned to the current bucket and designate the top $\tau_\text{local}$ (we set $\tau_\text{local}=20\%$  here) as local heads. For these heads, the attention mass on the current bucket is typically above 0.85. To verify their reliance on local information, we remove all KV caches associated with the identified local heads while keeping all other caches unchanged. As shown in Tab.~\ref{tab:local}, the Vbench~\cite{huang2023vbench} score indicates that this compression strategy incurs only a minor performance drop compared to the original Self Forcing, whereas removing the KV cache from an equivalent number of non-local heads leads to substantial degradation.

\noindent\textbf{\emph{Anchor Heads.}}
These heads exhibit significantly elevated first-frame attention as shown in Fig.~\ref{fig:attn_map} and Fig.~\ref{fig:headrole} (a). 
Based on Fig.~\ref{fig:headrole} (a), we identify anchor heads using the following criterion: attention heads whose attention weight to the first latent frame is in the top $\alpha_{\text{anchor}}$ percentile are designated as anchor heads (we set $\alpha_{\text{anchor}}=25 \%$ here).
We verify their role by removing the sink frame from the KV cache of these heads only. As shown in Fig.~\ref{fig:anchor}, evicting this frame from anchor heads causes rapid quality degradation, whereas the same operation on other heads produces no visible degradation. This confirms that anchor heads stabilize generation, which we attribute to the distinctive statistics of the first latent frame: it encodes only the initial image without temporal compression~\cite{huang2025self}, serving as a stable long-range reference. 
We also observe that this anchoring behavior is tied to the sink frame itself rather than the first cache position: replacing it with any other frame at the same position does not elicit high attention.

\noindent\textbf{\emph{Memory Heads.}}
The remaining attention heads as shown in Fig.~\ref{fig:attn_map} all distribute their attention \textbf{broadly across the entire range of historical frames}, effectively leveraging the full past context. These heads serve to aggregate long-range historical information and thus merit a larger KV cache budget to facilitate better long-term memory.

\noindent\textbf{Stability of Profiling Results.} We present whole profiling results in Fig.~\ref{fig:headrole} (b) for all attention heads. We further observe that the profiling results remain stable across different AR steps, denoising steps, and prompts with detailed validation process, computational overhead and  results provided in the \emph{supplementary}.
This stability suggests that the identified head categorization reflects an intrinsic structural property of the model.

\subsection{Head-wise KV Cache Allocation.}
Based on above insights, we propose a head-wise KV cache allocation framework on pretrained AR video DiTs, enabling robust minute-level video generation with long-horizon consistency and prompt-guided interactive long video generation.

\noindent\textbf{Local Heads \& Anchor Heads with KV Cache Pruning.}
Since \textbf{local heads} attend exclusively to local regions for detail refinement and motion continuity, we retain only the nearest neighboring frame and the current block, i.e., $K_{(3i{-}4)s:3is}$ for the $i$-th block.
\textbf{Anchor heads} exhibit unique sensitivity to the first latent frame, which possesses distinct statistical properties and serves as an anchor frame. As observed in Fig.~\ref{fig:headrole}, these heads also allocate partial attention to tokens in the middle bucket. To improve robustness against potential misclassification at the category boundary, we retain the first three frames as anchor frames alongside the same local tokens used by local heads, i.e., $[K_{0:3s}, K_{(3i{-}4)s:3is}]$.
Both strategies free substantial cache budget for memory heads while removing redundant historical tokens that may introduce noise path.

\noindent\textbf{Memory Heads with Hierarchical Memory System.} 
Unlike local and anchor heads, memory heads attend broadly across the entire history to aggregate long-range context. Simply applying a sliding window~\cite{yang2025longlive} discards intermediate context, causing narrative inconsistency and subject drift. We therefore introduce a hierarchical memory with two complementary tiers: \textbf{fast memory} $\mathcal{M}_{\text{fast}}$ (with size of $B_{\text{fast}}$) for the immediate temporal neighborhood, and \textbf{episodic memory} $\mathcal{M}_{\text{epi}}$ (with size of $B_{\text{epi}}$) storing representative KV cache from earlier segments.
$\mathcal{M}_{\text{fast}}$ retains KV cache from the most recent $B_{\text{fast}}$ frames preceding the current block, i.e., $ K_\text{$\mathcal{M}_{\text{fast}}$ }=K_{(3i-B_{\text{fast}}-3)s:(3i-3)s}$, providing fine-grained short-term memory for motion continuity and visual coherence. It is updated via FIFO policy at each AR step.

\noindent\textbf{Dynamic Update for Episodic Memory.}
The episodic memory $\mathcal{M}_{\text{epi}}$ stores a compact set of representative KV cache entries from earlier generated frames, enabling memory heads to recall distant scene elements and maintain long-term narrative consistency. Its design is guided by two principles: the stored KV cache (1) should be as clean as possible, with minimal error accumulation, and (2) should capture novel scenes and identities not yet represented in memory.

Prior work has shown that KV cache associated with new scenes exhibits less error accumulation compared to frames that are repeatedly conditioned on a long history within the same scene~\cite{yang2026stableworld}. Inspired by it, we adopt \textbf{scene novelty} as the episodic memory admission criterion, as new scenes carry both cleaner KV cache and previously unseen content.
When a block exits the fast memory window, we select its first frame as the representative candidate, since consecutive frames within a short chunk exhibit high temporal redundancy~\cite{ji2025memflow}. Scene novelty is measured via key embedding similarity, motivated by findings that key embeddings effectively capture temporal redundancy across frames~\cite{kim2025infinipot} with more validation in the \emph{supplementary}.

Specifically, assume the episodic memory currently contains $N_{\text{epi}}$ stored
frames at each attention layer $l \in \{1, \ldots, L\}$, where $L$ is the total
number of layers. To ensure semantic continuity, we enforce a consistent admission
policy across all layers, i.e., the same frame is either admitted to or excluded
from episodic memory at every layer simultaneously. We therefore aggregate
novelty score across layers when evaluating scene novelty.
We first apply mean pooling along the token dimension to compress the key cache of the candidate frame and each episodic entry into compact representatives
$K_c^{(l,h)} \in \mathbb{R}^{1 \times d}$ and
$K_{\text{epi},j}^{(l,h)} \in \mathbb{R}^{1 \times d}$, where
$j \in \{1, \ldots, N_{\text{epi}}\}$ indexes the episodic entries and
$h \in H_{\text{mem}}$ indexes the memory heads within each layer. Such compressed key representations have been shown to be sufficiently expressive for capturing frame-level semantics in generation tasks~\cite{cai_mixture_2025, ji2025memflow}. The novelty score is the maximum average cosine similarity to all episodic entries across layers and memory heads:
\begin{equation}
\label{eq:novelty}
\small
    \delta(K_c) = \max_{j \in \{1,\ldots,N_{\text{epi}}\}}
    \frac{1}{|H_{\text{mem}}| \times L}
    \sum_{l=1}^{L} \sum_{h \in H_{\text{mem}}}
    \mathrm{cossim}\!\left(K_c^{(l,h)},\, K_{\text{epi},j}^{(l,h)}\right),
\end{equation}
where $|H_{\text{mem}}|$ is the total number of memory heads. A low $\delta(K_c)$ indicates the candidate is visually distinct from stored entries. The candidate is admitted into the episodic memory only if $\delta(K_c) < \tau_{\text{novel}}$. Notably, the novelty score computation can be parallelized across layers and heads, introducing negligible computational overhead.

\noindent\textbf{Prompt-guided Compression.}
When the episodic memory reaches its full capacity $B_{\text{epi}}$, rather than directly discarding entries, we compress redundant entries into a dedicated \emph{summary frame} $\mathcal{S}$ via token-wise importance selection. The summary frame is placed at the beginning of the episodic memory.

Specifically, for the $l$-th layer, on the initial overflow, no summary frame exists yet.  We identify the most redundant pair $(\hat K_{\text{epi},i}^{(l)}, \hat K_{\text{epi},j}^{(l)})\in \mathbb{R}^{|H_{\text{mem}}|\times s \times d}$, $i, j \in \{ 1, \ldots, B_{\text{epi}}\}$, as the pair with highest pairwise key-embedding cosine similarity. Note that all pairwise cosine similarities have already been computed during the scene novelty evaluation in Eq.~\ref{eq:novelty}, incurring no additional cost. The selected pair is
then compressed to initialize $\mathcal{S}$. On subsequent overflows, we select the non-summary frame $\hat K_{\text{epi},j}^{(l)}$, $j\in\{2,\cdots,B_\text{epi}\}$, with highest average similarity to its neighboring entries and merge it into $\mathcal{S}$.

For the compression procedure, we operate at the spatial token granularity. Concatenating KV caches of $K_\mathcal{S}^{(l)}$ and $\hat K_{\text{epi,}j}^{(l)}$ (or $\hat K_{\text{epi},i}^{(l)}$ and $\hat K_{\text{epi},j
}^{(l)}$ for initial case) along token dimension yields $2s$ key and value tokens, i.e., $K_\text{cat}^{(l)}=[K_\mathcal{S}^{(l)}, \hat K_{\text{epi},j}^{(l)}]$ and $V_\text{cat}^{(l)}=[V_\mathcal{S}^{(l)}, \hat V_{\text{epi},j}^{(l)}]$ ($K_\text{cat}^{(l)}=[\hat K_{\text{epi},i}^{(l)}, \hat K_{\text{epi},j}^{(l)}]$, $V_\text{cat}^{(l)}=[\hat V_{\text{epi},i}^{(l)}, \hat V_{\text{epi},j}^{(l)}]$ for initial case), where $K_{\text{cat}}^{(l)}, V_{\text{cat}}^{(l)} \in \mathbb{R}^{|H_{\text{mem}}|\times2s \times d}$.
Candidate tokens are scored by average cosine similarity to the current textual key embedding across memory heads: $r^{(l)} = \frac{1}{|H_{\text{mem}}|}\sum_{h\in H_{\text{mem}}}\operatorname{cossim}(K_\text{cat}^{(l,h)},\, K_{\text{prompt}}^{(l,h)})$
where $K_{\text{prompt}}^{(l,h)} \in \mathbb{R}^{1 \times d}$ is the current textual key per head, and $r^{(l)} \in \mathbb{R}^{2s}$ is the resulting textual alignment score for each token.
The top-$s$ tokens, ranked by $r^{(l)}$, are selected to form the updated summary frame $\mathcal{S}$: $(K_\mathcal{S}^{(l)},V_\mathcal{S}^{(l)}) \leftarrow \operatorname{Top-}s\!\left((K_\text{cat}^{(l)}, V_\text{cat}^{(l)}),\; r^{(l)} \right)$. In the multi-prompt setting, scoring with the active prompt embedding in
$r^{(l)}$ naturally biases token retention toward content that is semantically relevant to the current prompt as it evolves, which is crucial for maintaining narrative coherence across prompt transitions. After the above process, the final key sequence assembled for each memory head is
$[K_{\mathcal{M}_{\text{epi}}},\; K_{\mathcal{M}_{\text{fast}}},\;
K_{(3i{-}3)s:3is}]$.

\subsection{Head-wise RoPE Re-encoding}
In AR video generation, global frame indices as RoPE coordinates grow linearly with generated blocks, while cached anchor and episodic memory frames retain outdated indices from their original steps. The resulting relative positions between queries and cached keys may exceed the pretrained range, producing positional O.O.D. analogous to LLMs~\cite{jin2024llm}, which exacerbates error accumulation and flickering artifacts.
To address this, we propose head-wise RoPE re-encoding. Rather than storing position-encoded KV cache and subsequently correcting temporal embeddings~\cite{yi2025deep}, we cache keys and values with only spatial-dimension RoPE applied, omitting frame-level temporal encoding. This preserves spatial positional information of cached tokens—particularly important for importance-compressed tokens whose spatial semantics remain intact regardless of temporal reassignment. After each cache update, we apply parallelizable head-wise re-encoding of frame-level indices to the assembled key sequences across all heads, simplifying the overall procedure.
Specifically, at the $i$-th block with $f{=}3$ frames, each head $h$ operates over a distinct assembled key sequence of head-specific length. Let $F^{(h)}_i$ denote the total cached frames (including the current block) for head $h$, and $K^{(h)}_i$ the concatenated key sequence for attention. We assign contiguous frame indices $\{0, 1, \ldots, F^{(h)}_i{-}1\}$ to $K^{(h)}_i$, and indices starting from $F^{(h)}_i - f$ to $Q_i^{(h)}$, leaving spatial (height, width) RoPE components unchanged:
\begin{equation}
\label{eq:rope}
    Q_i^{(h)}:\;\{F^{(h)}_i{-}f,\;\ldots,\;F^{(h)}_i{-}1\},\qquad
    K^{(h)}_i:\;\{0,\;\ldots,\;F^{(h)}_i{-}1\}
    \small
\end{equation}
This aligns each query with the last $f$ frames of its key sequence. Since $F^{(h)}_i$ is bounded by the cache capacity of each head, all temporal relative positions remain within the pretrained range regardless of generation length.

\subsection{Implementation With Efficiency Optimization}
Head-wise KV cache allocation improves long video generation quality but introduces computational challenges due to variable-sized caches across heads. We address this from two complementary directions.

\noindent\textbf{Variable-Length FlashAttention.}
The variable-length FlashAttention interface~\cite{dao2022flashattention, dao2023flashattention}, originally designed for continuous batching, naturally accommodates our head-wise scheme. We treat each head as an independent sequence and invoke a single $\texttt{flash\_attn\_varlen\_func}$ call for all heads simultaneously, avoiding per-head kernel launches.

\noindent\textbf{Fused Triton Kernels.}
To reduce head-wise cache management overhead, we implement fused Triton~\cite{tillet2019triton} kernels. Two kernels handle KV cache updates: one for rolling the cache across all heads, another for updating episodic memory entries of $H_{\text{mem}}$. After update, each head assembles its sequence by concatenating its sink region, episodic memory (for $H_{\text{mem}}$ only), and local region, then applies RoPE re-encoding (Eq.~\ref{eq:rope}) before packing into a flat buffer for variable-length attention. We fuse these three operations into a single Triton kernel. Together, these optimizations ensure negligible throughput overhead compared to the uniform-cache baseline.

%% file: sections/experiments.tex
\section{Experiments}
\label{sec:exp}
\subsection{Experimental Settings}
\noindent\textbf{Implementation details.} We implement our method on Self Forcing~\cite{huang2025self}, a causal four-step generator distilled from Wan2.1-T2V-1.3B~\cite{wan2025} that natively produces five-second videos at 16 FPS with a resolution of $832\times480$. Each autoregressive step generates a block of $f=3$ latent frames. For memory system, we set $B_\text{epi}=5$, $B_\text{fast}=3$ and $\tau_\text{novel}=0.95$ and we update episode memory every three generation intervals.

\noindent\textbf{Baselines.}
We compare Head Forcing against representative baselines spanning both training-based and training-free paradigms. Training-based methods include Self Forcing~\cite{huang2025self}, CausVid~\cite{yin2025slow}, Rolling Forcing~\cite{liu2025rolling}, and LongLive~\cite{yang2025longlive}. Training-free methods include Deep Forcing~\cite{yi2025deep} and Infinity-RoPE~\cite{yesiltepe2025infinity}. For the multi-prompt interactive generation setting, we compare with Self Forcing~\cite{huang2025self}, LongLive~\cite{yang2025longlive} and Infinity-RoPE~\cite{yesiltepe2025infinity}.

\noindent\textbf{Evaluation.}
We evaluate on VBench~\cite{huang2023vbench,huang2025vbench++} following prior protocols~\cite{yi2025deep, yesiltepe2025infinity, yang2025longlive,huang2025self}. For one-prompt long video generation, we sample 100 prompts from MovieGenBench~\cite{polyak2024movie} and generate videos at 30 and 60 seconds. We report common VBench-Long metrics and throughput (FPS). For multi-prompt interactive generation, we design 50 multi-prompt sequences following Longlive~\cite{yang2025longlive}, each with six successive prompts describing distinct actions or scene transitions, and evaluate on 60-second videos. Following prior work~\cite{yang2025longlive, ji2025memflow}, we report Quality Score, Consistency Score, and Aesthetic Score from VBench-Long~\cite{huang2025vbench++}, along with segment-wise CLIP scores at each prompt interval for text alignment.

\subsection{Quantitative Results}

\noindent\textbf{Long Video Generation.}
Tab.~\ref{tab:vbench_long} presents the quantitative comparison for 30 second and 60 second video generation using one prompt. Head Forcing consistently outperforms both training-free across nearly all metrics. Unlike methods relying on uniform attention sinks that often induce motion stagnation, our head-wise KV cache framework effectively maintains long-horizon consistency and perceptual quality while preserves motion richness.
At the same time, our throughput is comparable to baselines, confirming that the head-wise allocation combined with efficiency optimization introduces negligible overhead. 

\begin{table*}[t]
    \centering
    \caption{\textbf{Quantitative comparison on long video generation.} We evaluate Head Forcing against open-source AR video diffusion and training-free AR long video generation baselines on 30 s and 60 s videos across multiple quality metrics on VBench.}
    \label{tab:vbench_long}
    \setlength{\tabcolsep}{3.2pt}
    \renewcommand{\arraystretch}{1.1}
    \small
    \resizebox{\textwidth}{!}{%
    \begin{tabular}{lcccccccc}
        \toprule
        \textbf{Model} &
        \shortstack{\textbf{Throughput}\\\textbf{(FPS)$\uparrow$}} &
        \shortstack{\textbf{Dynamic}\\\textbf{Degree$\uparrow$}} &
        \shortstack{\textbf{Motion}\\\textbf{Smoothness$\uparrow$}} &
        \shortstack{\textbf{Temporal}\\\textbf{Flickering$\uparrow$}} &
        \shortstack{\textbf{Imaging}\\\textbf{Quality$\uparrow$}} &
        \shortstack{\textbf{Aesthetic}\\\textbf{Quality$\uparrow$}} &
        \shortstack{\textbf{Subject}\\\textbf{Consistency$\uparrow$}} &
        \shortstack{\textbf{Background}\\\textbf{Consistency$\uparrow$}} \\
        \midrule
        \rowcolor{gray!7} \textit{Training-based} & \multicolumn{8}{c}{\textbf{30 seconds}}\\
        \cmidrule(lr){2-9}
        Rolling Forcing~\cite{liu2025rolling} & 15.99 & 33.92 & \underline{98.80} & 98.62 & \underline{70.21} & 61.26 & \textbf{98.12 }& 97.01 \\
        LongLive~\cite{yang2025longlive} & 19.03 & \underline{40.72}& \textbf{98.83} &\textbf{ 98.80}& 69.22 & \underline{61.44} & 97.97 &\textbf{ 97.15} \\
        CausVid~\cite{yin2025slow} & 15.83 & 32.21 & 98.21 & 98.47 & 65.78 & 59.87 & 97.64 & 96.84 \\
        Self Forcing~\cite{huang2025self} & 15.83 & 35.22 & 98.42 & 98.58 & 68.24 & 60.16 & 97.62 & 96.77 \\
        \rowcolor{gray!7} \multicolumn{9}{l}{\textit{Training-free}}\\
        Deep Forcing~\cite{yi2025deep} & 15.74 & 36.42 & 98.31 & 98.59& 69.33 & 60.70 & 97.85& 96.95 \\
        Infinity-RoPE~\cite{yesiltepe2025infinity} & 16.63 &38.27 & 98.64 & \underline{98.79}& 69.16 & 60.82 & 97.77 & 97.06 \\
        \rowcolor{blue!5} \textbf{Head Forcing (Ours)} & 15.81 & \textbf{42.14} & 98.76 &98.78 & \textbf{70.30}& \textbf{61.68} & \underline{98.07} & \underline{97.08} \\
        \midrule
        \rowcolor{gray!7} \textit{Training-based} & \multicolumn{8}{c}{\textbf{60 seconds}}\\
        \cmidrule(lr){2-9}
        Rolling Forcing~\cite{liu2025rolling} & 15.99 & 32.86& \underline{98.69}&98.57 & \textbf{70.09} & 60.97 & \textbf{97.87} & 96.87 \\
        LongLive~\cite{yang2025longlive} & 19.03 & \underline{40.29} & \textbf{98.75} & \underline{98.70} & 69.08 & \underline{61.11} & 97.72 & \textbf{96.92}\\
        CausVid~\cite{yin2025slow} & 15.83 & 31.08 & 98.26& 98.49 & 65.36 & 59.32 & 97.53 & 96.66\\
        Self Forcing~\cite{huang2025self} & 15.83  & 31.92 & 98.21 & 98.56 & 67.33 & 57.17 & 97.32& 96.53\\
        \rowcolor{gray!7}\multicolumn{9}{l}{\textit{Training-free}}\\
        Deep Forcing~\cite{yi2025deep} & 15.74 &35.75 & 98.38 & 98.63 & 68.93 & 60.39 & 97.70 & 96.82 \\
        Infinity-RoPE~\cite{yesiltepe2025infinity} & 16.63 & 37.19& 98.49 & 98.61 & 68.68 & 59.65 & 97.73 & 96.88 \\
        \rowcolor{blue!7} \textbf{Head Forcing (Ours)} & 15.81 & \textbf{41.37} & 98.67 & \textbf{98.72} & \underline{69.27}& \textbf{61.36} & \underline{97.81} & \underline{96.90 } \\
        \bottomrule
    \end{tabular}
    }
    \vspace{-10pt}
\end{table*}

\noindent\textbf{Prompt-Guided Interactive Video Generation.}
Tab.~\ref{tab:multi_prompt_60s} reports results under the multi-prompt 60-second setting, where prompts switch at 10-second intervals. Head Forcing achieves the best Aesthetic and Consistency Scores while remaining competitive in Quality Score. Notably, our method yields better segment CLIP scores after multiple prompt switches, indicating improved prompt alignment and contextual preservation.
Additionally, we conducted a user
study to compare with the aforementioned models in \textit{supplementary}.

\begin{table*}[t!]
    \centering
    \caption{\textbf{Quantitative comparison under multi-prompt 60 second setting} with representative prompt-guided interactive video models. All scores are measured over the whole sequence, except for the CLIP score.}
    \label{tab:multi_prompt_60s}
    \setlength{\tabcolsep}{4.2pt}
    \renewcommand{\arraystretch}{1.12}
    \small
    \resizebox{\textwidth}{!}{%
    \begin{tabular}{lccc|cccccc}
        \toprule
        \textbf{Method} &
        \shortstack{\textbf{Quality}}&
        \shortstack{\textbf{Consistency}}&
        \shortstack{\textbf{Aesthetic}}&
        \multicolumn{6}{c}{\textbf{CLIP Score}$\uparrow$} \\
        \cmidrule(lr){5-10}
        & \textbf{Score}$\uparrow$& \textbf{Score}$\uparrow$& \textbf{Score}$\uparrow$&
        \textbf{0--10\,s} &
        \textbf{10--20\,s} &
        \textbf{20--30\,s} &
        \textbf{30--40\,s} &
        \textbf{40--50\,s} &
        \textbf{50--60\,s} \\
        \midrule
        Self Forcing~\cite{huang2025self} & 83.52& 95.74 & 57.45& 26.25& 24.93& 23.46 & 21.92 & 21.82& 21.02\\
        LongLive~\cite{yang2025longlive} & \textbf{84.83}& 96.05 & \underline{60.29}& \textbf{26.63} & \textbf{25.77} & \textbf{24.75}& \underline{24.19}& \underline{24.52} & \underline{24.11} \\
        Infinity-RoPE~\cite{yesiltepe2025infinity} & 84.66& 96.10& 59.33& \underline{26.78}& \underline{25.43}& 24.38& 23.55& 23.82& 23.53\\
        \rowcolor{blue!5} \textbf{Head Forcing(Ours)} & \underline{84.74}& \textbf{96.69}& \textbf{60.67}& 26.54& 25.36& \underline{24.64}& \textbf{24.27}& \textbf{24.62}& \textbf{24.22} \\
        \bottomrule
    \end{tabular}
    }
    \vspace{-15pt}
\end{table*}

\subsection{Qualitative Results}
\begin{figure}
\vspace{-10pt}
    \centering
    \includegraphics[width=1\linewidth]{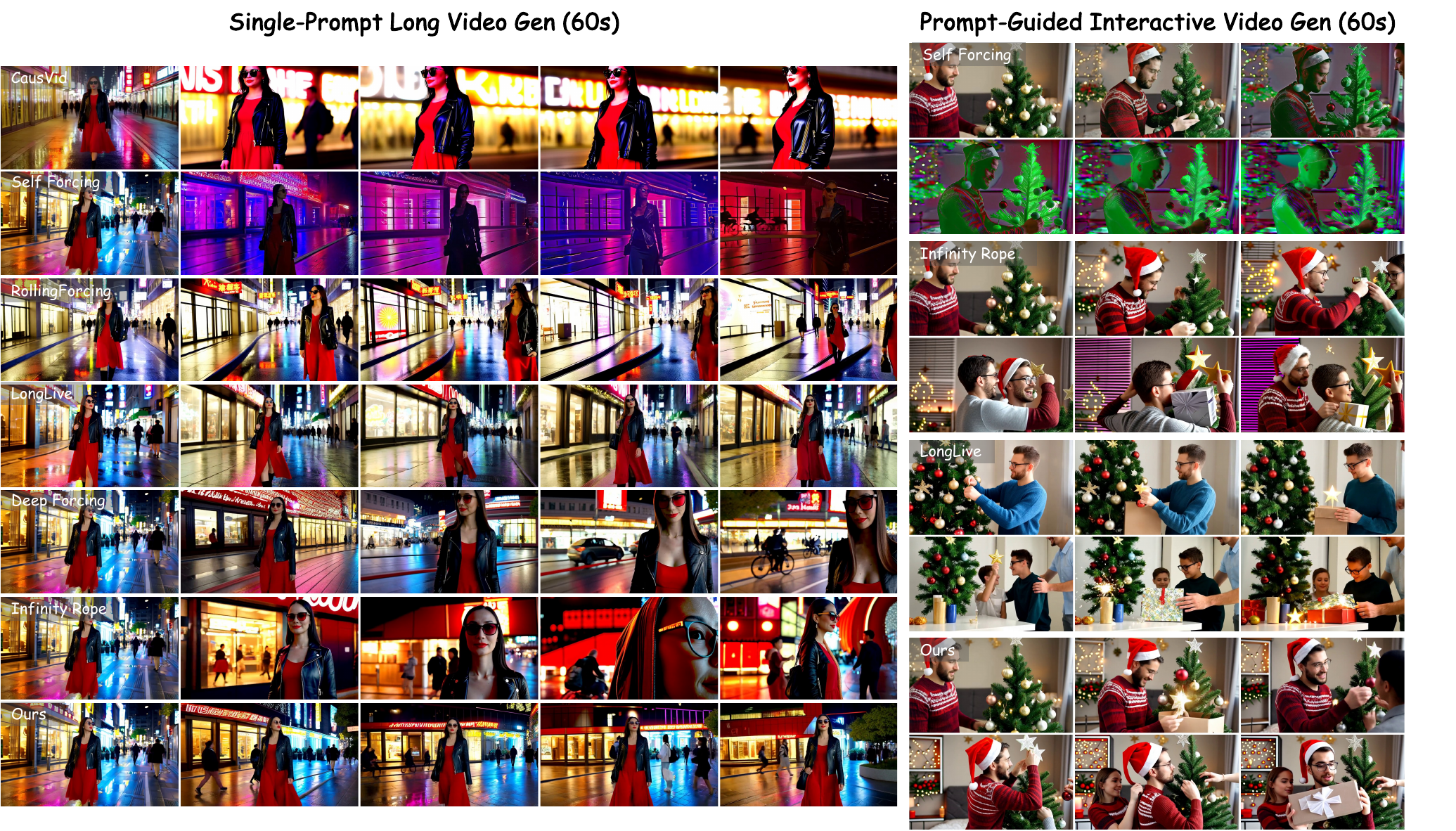}
    \caption{\textbf{Qualitative Results.} Qualitative comparison on 60\,s single-prompt long video generation and 60\,s prompt-guided interactive generation. }
    \label{fig:quality}
    \vspace{-12pt}
\end{figure}

\noindent\textbf{Long Video Generation.}
Fig.~\ref{fig:quality} presents qualitative comparisons for 60-second generation. Self Forcing~\cite{huang2025self} and CausVid~\cite{yin2025slow} exhibit progressive color drift and oversaturation beyond their native 5 second training horizon. Deep Forcing~\cite{yi2025deep} and Infinity-RoPE~\cite{yesiltepe2025infinity} slightly mitigate quality degradation but still suffer from reduced motion dynamics. Rolling Forcing~\cite{liu2025rolling} and LongLive~\cite{yang2025longlive} preserve visual stability through trained attention sinks but tend to produce repetitive content with limited scene evolution. In contrast, Head Forcing maintains high visual fidelity, subject identity, and scene dynamics throughout the entire generation.

\noindent\textbf{Prompt-Guided Interactive Video Generation.}
Fig.~\ref{fig:quality} also illustrates multi-prompt interactive generation. Given six successive prompts describing different actions and scenes, Head Forcing produces smooth transitions while faithfully following each instruction, maintaining subject identity and scene coherence across prompt. In comparison, LongLive~\cite{yang2025longlive} and Infinity-RoPE~\cite{yesiltepe2025infinity} achieve reasonable prompt adherence but exhibit identity drift over extended sequences due to limited context while Self Forcing~\cite{huang2025self} exhibits more pronounced error accumulation.

\subsection{Ablation Study}
\label{sec:ablation}

We conduct ablation studies on 60\,s long video generation and  multi-prompt interactive generation to evaluate the contribution of each component and validate key design choices.
More ablation studies are provided in the \emph{supplementary}.

\begin{table}[t]
\centering
\caption{\textbf{Component-wise ablation on 60\,s generation.}
HW = head-wise KV cache allocation, HM = hierarchical memory, RoPE = head-wise RoPE re-encoding.}
\label{tab:component_ablation}
\resizebox{\linewidth}{!}{%
\begin{tabular}{l ccc cccc ccc}
\toprule
& \multicolumn{3}{c}{\textbf{Modules}}
& \multicolumn{4}{c}{\textbf{60\,s} \ (\textbf{Single Prompt Long Video Generation})}
& \multicolumn{3}{c}{\textbf{60\,s} \ (\textbf {Multi-Prompt})} \\
\cmidrule(lr){2-4} \cmidrule(lr){5-8} \cmidrule(lr){9-11}
& Head-Wise & Hierarchical & RoPE
& Dynamic & Motion & Subject 
& Imaging& Quality& Consistency & Clip Score\\
& Allocation & Memory & Re-encoding
& Degree$\uparrow$ & Smoothness$\uparrow$ & Consistency$\uparrow$ 
& Quality$\uparrow$& Score$\uparrow$& Score$\uparrow$& Average$\uparrow$\\
\midrule
(A) Baseline   & \xmark & \xmark & \xmark &  31.92&  98.42&  97.62&  67.33&  83.52&  95.74&  23.20\\
(B) $+$ HW     & \cmark & \xmark & \xmark &  37.22&  98.49&  97.65&  68.14&  84.10&  96.33&  24.31\\
(C) $+$ HM     & \cmark & \cmark & \xmark &  39.09&  98.54&  97.73&  68.93&  84.34&  96.51&  24.61\\
\rowcolor{blue!7}(D) Ours   & \cmark & \cmark & \cmark &  \textbf{41.37}&  \textbf{98.69}&  \textbf{97.81}&  \textbf{69.27}&  \textbf{84.74}&  \textbf{96.69}&  \textbf{24.95}\\
\bottomrule
\end{tabular}}
\vspace{-15pt}
\end{table}

\noindent\textbf{Impact of Main Components.}
Tab.~\ref{tab:component_ablation} isolates the contribution of each main component by progressively adding them to the (A) Self-Forcing~\cite{huang2025self} baseline. (B) Introducing head-wise KV cache allocation alone (where memory heads still use a sliding window as in Self-Forcing under the same KV cache budget) improves quality, consistency, and dynamic degree, as pruning irrelevant context from local heads and stabilizing generation through anchor frames while allowing memory heads to adopt a larger context window with the saved budget.
(C) Adding the hierarchical memory system for memory heads further boosts contextual consistency.
(D) Finally, enabling RoPE re-encoding yields additional gains in motion smoothness and visual quality.
These results demonstrate that three main components all contribute to our methods’ outstanding performance.

\begin{wrapfigure}{rt}{0.75\textwidth}
    \vspace{-.85cm}
    \centering
    \includegraphics[width=\linewidth]{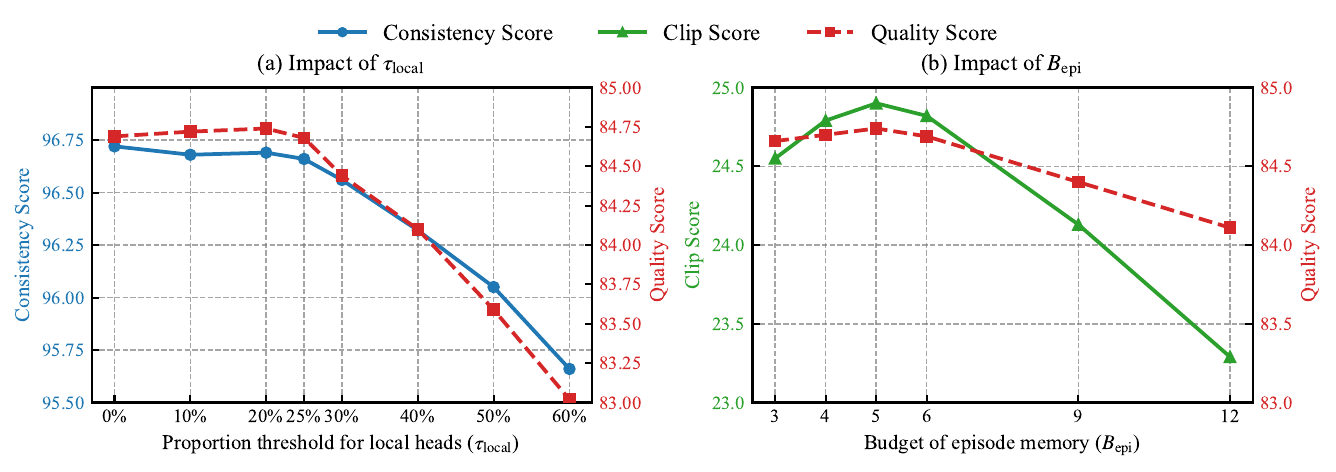}
    \caption{
    Hyperparameter Analysis of $\tau_{\text{local}}$ and $B_{\text{epi}}$.
    }
    \label{fig:hyper}
    \vspace{-2.2em}
\end{wrapfigure}
\noindent\textbf{Detailed Validation of Head-wise Strategy.}
To further validate whether the improvement stems from the head-wise  KV cache allocation, we compare four strategies under the same total KV cache budget: (1) Random Allocation, randomly dividing the same proportion of heads into three categories.  (3) Local $+$ Memory, merging anchor heads into memory heads. (4) Anchor $+$ Memory, merging local heads into memory heads. (4) Anchor $+$ Memory, merging local heads into memory heads. 
As shown in Tab.~\ref{tab:headwise},  Uniform Memory, Anchor $+$ Memory, and Local $+$ Memory waste cache budget and introduce redundant noise paths, degrading efficiency and  reducing visual quality.

\begin{table}[t]
\centering
\caption{\textbf{Detail validation of head-wise strategy.}}
\vspace{-10pt}
\label{tab:headwise}
\scriptsize
\resizebox{\textwidth}{!}{%
\begin{tabular}{l cccc ccc}
\toprule
& \multicolumn{4}{c}{\textbf{60\,s} \ (\textbf{Single Prompt Long Video Generation})}
& \multicolumn{3}{c}{\textbf{60\,s} \ (\textbf{Multi-Prompt})} \\
\cmidrule(lr){2-4} \cmidrule(lr){2-5} \cmidrule(lr){6-8}
\textbf{Strategy} & Dynamic & Motion & Subject 
& Imaging& Quality& Consistency& Clip Score\\
& Degree$\uparrow$ & Smoothness$\uparrow$ & Consistency$\uparrow$ 
& Quality$\uparrow$& Score$\uparrow$& Score$\uparrow$& Average$\uparrow$\\
\midrule
Random Allocation  &  32.21&  98.29&  97.38&  67.92&  83.55&  95.88&  23.55\\
Uniform Memory  &  36.73&  98.41&  97.53&  68.12&  84.21&  96.11&  24.21\\
Local $+$ Memory     &  34.28&  98.47&  97.56&  67.98&  84.35&  96.19&  24.12\\
Anchor $+$ Memory     &  39.73&  98.55&  97.70&  68.86&  84.66&  96.54&  24.65\\
\rowcolor{blue!7}\textbf{Head Forcing}  &  \textbf{41.37}&  \textbf{98.69}&  \textbf{97.81}&  \textbf{69.27}&  \textbf{84.74}&  \textbf{96.69}&  \textbf{24.95}\\
\bottomrule
\end{tabular}}
\vspace{-12pt}
\end{table}

\noindent\textbf{Episodic Memory Design.}
(1) Impact of different dynamic update strategies:
For dynamic update strategy, we compare our scene novelty based criterion against uniform temporal sampling which admits frames into episodic memory at fixed intervals.
As shown in Tab.~\ref{tab:episodic_ablation}, 
Our novelty-based criterion selectively retains frames from distinct scenes, yielding the best scene consistency and visual quality.
(2) Impact of prompt-guided compression: We compare our compression method with FIFO eviction and random selection which randomly samples $s$ tokens along the token dimension from candidate frames.
Our prompt-guided method preserves diverse scene coverage, and improves average CLIP scores in the multi-prompt setting by biasing token retention toward currently relevant content.
\begin{table}[t]
\centering
\caption{\textbf{Ablation on episodic memory design.} Top: admission criterion. Bottom: compression strategy upon overflow.}
\label{tab:episodic_ablation}
\vspace{-5pt}
\setlength{\tabcolsep}{4pt}
\resizebox{\textwidth}{!}{%
\begin{tabular}{l cccc ccc}\toprule

& \multicolumn{4}{c}{\textbf{60\,s} \ (\textbf{Single Prompt Long Video Generation})}
& \multicolumn{3}{c}{\textbf{60\,s} \ (\textbf{Multi-Prompt})} \\
\cmidrule(lr){2-4} \cmidrule(lr){2-5} \cmidrule(lr){6-8}
\textbf{Strategy} & Dynamic & Motion & Subject 
& Imaging& Quality& Consistency& Clip Score\\
& Degree$\uparrow$ & Smoothness$\uparrow$ & Consistency$\uparrow$ 
& Quality$\uparrow$& Score$\uparrow$& Score$\uparrow$& Average$\uparrow$\\
\midrule
\multicolumn{8}{c}{\emph{Admission Criterion}}\\
 \cmidrule(lr){2-8}
Uniform sampling        & 37.42& 98.53& 97.71& 68.87& 84.61& 96.21&24.59\\
\rowcolor{blue!7}Novelty-based (Ours)     & \textbf{41.37}& \textbf{98.69}& \textbf{97.81}& \textbf{69.27}& \textbf{84.74}& \textbf{96.69}&\textbf{24.95}\\
\midrule
\multicolumn{8}{c}{\emph{Compression Strategy}}\\
\cmidrule(lr){2-8}

FIFO eviction                          & 39.42& 98.62& 97.73& 69.04& 84.68& 96.52& 24.71\\
Random selection& 40.42& 98.59& 97.76& 68.98&  84.64& 96.61& 24.68\\
\rowcolor{blue!7}Ours    & \textbf{41.37}& \textbf{98.69}& \textbf{97.81}& \textbf{69.27}& \textbf{84.74}& \textbf{96.69}& \textbf{24.95}\\ 
\midrule

\end{tabular}}
\vspace{-16pt}
\end{table}

\noindent\textbf{Hyperparameter Analysis.} (1) Impact of episode memory budget $B_\text{epi}$ and proportion threshold $\tau_{\text{local}}$ for local head: Results are reported in Fig.~\ref{fig:hyper} that performance is stable across a reasonable range for the two parameters, indicating that the method does not require careful tuning. 
However, an overly small $\tau_{\text{local}}$ which assigns an excessive number of local heads, degrades generation quality. 
And an overly large $B_{\text{epi}}$ 
 may also harm performance, as the proportion of episodic memory tokens significantly outweighs that of the local window.
(2) We put other hyperparameter evaluation in the \emph{supplementary}.

%% file: sections/conclusion.tex
\section{Conclusion}
We presented Head Forcing, a training-free framework that leverages the functional heterogeneity of attention heads for long autoregressive video generation. By profiling heads into three distinct types, we assign tailored KV cache strategies to each, replacing uniform allocation with head-wise management. Memory heads are further equipped with a hierarchical memory system for long-range consistency, and a head-wise RoPE re-encoding scheme ensures positional encodings remain within the pretrained range. Without additional training, Head Forcing extends generation from 5 seconds to minute-level duration and consistently outperforms existing baselines on VBench.

%% file: sections/supp.tex

\section{Profiling Details, Results and Stability Validation}
\label{sec:supp_profiling}

This section provides comprehensive details on the attention head profiling procedure introduced in Sec.~3.2 of the main paper, including the profiling setup (Sec.~\ref{sec:supp_profiling_setup}), additional profiling results (Sec.~\ref{sec:supp_full_results}), extensive stability validation across AR steps, denoising timesteps, text prompts (Sec.~\ref{sec:supp_stability}) and computational overhead of profiling process (Sec.~\ref{sec:computational}).

\subsection{Detailed Profiling Setup}
\label{sec:supp_profiling_setup}

\noindent\textbf{Profiling Model and Rollout Setting.}
We profile Self Forcing~\cite{huang2025self}, a causal four-step generator distilled from Wan2.1-T2V-1.3B~\cite{wan2025}.
The underlying DiT consists of $L{=}30$ transformer layers, each containing $H{=}12$ self-attention heads with a per-head dimension of $d{=}128$, yielding $L \times H = 360$ attention heads in total.
Each autoregressive (AR) step generates a block of $f{=}3$ latent frames at a spatial resolution of $52 \times 30$ ($s{=}1560$ tokens per frame), conditioned on previously generated context via a KV cache of fixed window size $W{=}21$ frames combined with the sink frame under a FIFO eviction policy.

We profile the model during long-rollout generation on a prompt set $\mathcal{P}$ containing 20 prompts to identify each head's role for long video generation. For each prompt, we generate a rollout of $30$ seconds (about $42$ AR steps) and record self attention statistics at a randomly sampled set of autoregressive block indices $\mathcal{I}$ containing three AR steps. To avoid bias from the very early generation stage, where the historical context is still too short to exhibit stable long-range behavior, we begin profiling from block $3$. For each sampled block, we collect attention statistics from all denoising steps in the diffusion process.

Specifically, at every AR step $i$ and every denoising timestep $t \in \{0,1,2,3\}$, we record the full per-head attention map $A^{(l,h)} \in \mathbb{R}^{3s \times 3is}$ for all layers $l \in \{1,\ldots,30\}$ and heads $h \in \{1,\ldots,12\}$.
We then compute the three bucket-level attention proportions defined in Eq.~3 of the main paper:
the sink proportion $p_{\text{sink}}$, the middle proportion $p_{\text{middle}}$ , and the current proportion $p_{\text{current}}$.
Unless otherwise stated, the profiling statistics reported below are computed by first averaging over all denoising timesteps at each AR step, then averaging over randomly sampled three AR steps.
Based on the aggregated profiling statistics, heads are classified as follows:
\begin{itemize}
    \item \textbf{Anchor heads}: those whose $p_{\text{sink}}$ exceeds the top $\alpha_{\text{anchor}}{=}25 \%$ percentile are designated as anchor heads.
    \item \textbf{Local heads}:  among the remaining heads, the top $\tau_{\text{local}}{=}20\%$ of heads ranked by $p_{\text{current}}$. These heads typically exhibit $p_{\text{current}} > 0.85$.
    \item \textbf{Memory heads}: remaining heads, which distribute attention broadly across the full historical context.
\end{itemize}

\subsection{Additional Profiling Results and Analysis}
\label{sec:supp_full_results}

\noindent\textbf{Complete analysis of layer-wise head role distribution.}
Fig.~3\,(b) provides the full layer-wise head role distribution for all 30 layers.
Several patterns emerge:
\begin{itemize}
    \item Local heads are predominantly found in the \textbf{early layers} (layers 0--8), consistent with early layers handling low-level feature extraction and local texture synthesis.
    \item Anchor heads appear across \textbf{all layers} but with higher concentration in the early-to-middle layers (layers 0--14), reflecting that structural stabilization requires hierarchical anchoring at multiple representation levels.
    \item Memory heads are most prevalent in the \textbf{middle and deep layers} (layers 8--27), where high-level semantic aggregation over long temporal contexts is most critical.
\end{itemize}
This layer-wise distribution aligns with the well-established finding that transformer layers form a hierarchy from local features to global semantics, further supporting the functional interpretation of the three head types.

\begin{figure}[!htbp]
    \centering
    \includegraphics[width=.95\linewidth]{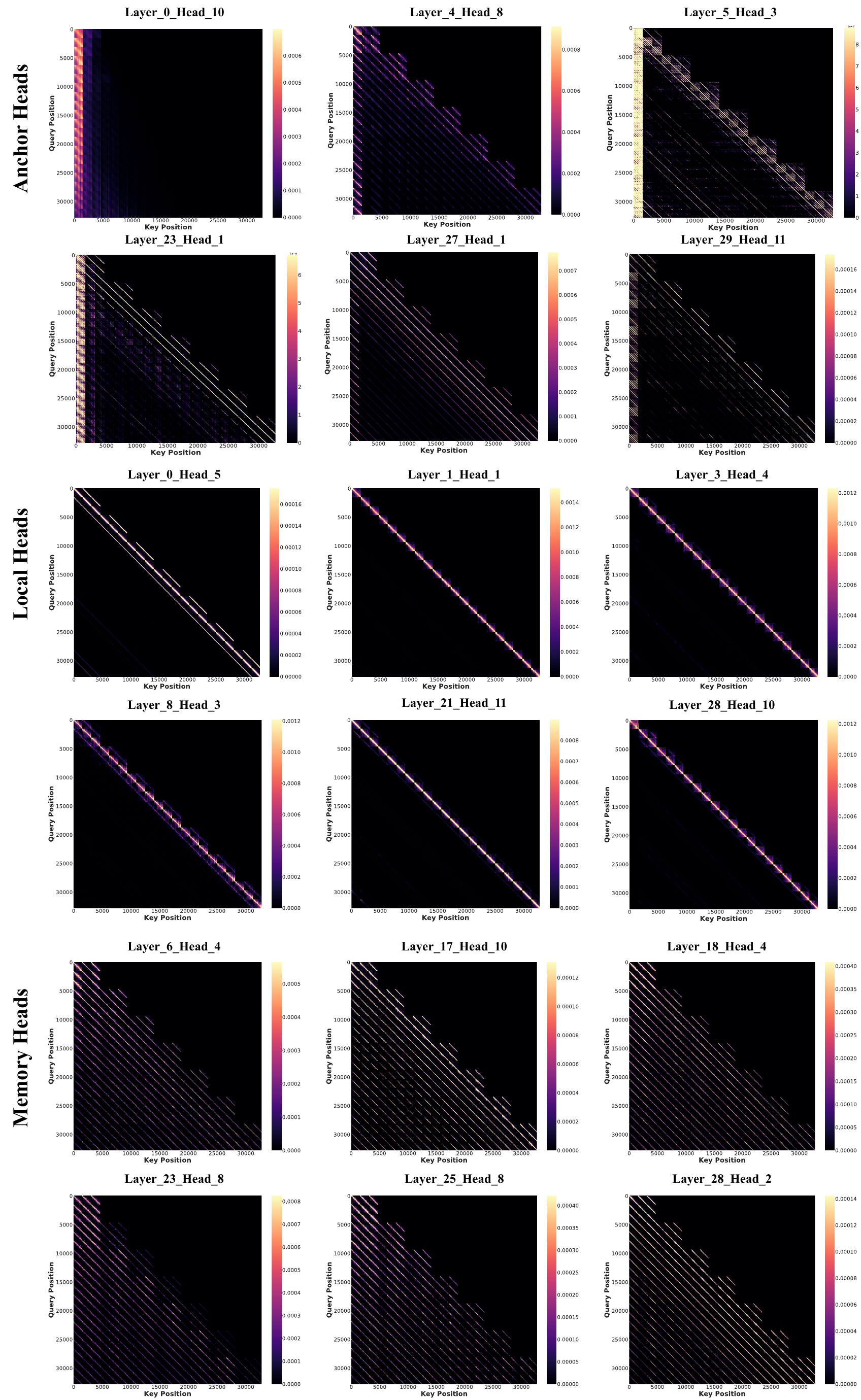}
    \caption{\textbf{More visualization of attention map.}}
    \label{fig:supp_attn_gallery}
\end{figure}

\noindent\textbf{More Visualization of Attention Map.}
Fig.~\ref{fig:supp_attn_gallery} presents an expanded gallery of per-head attention maps beyond the three examples shown in main paper Fig.~2.
We display more attention maps for representative heads from each type, sampled from different layers to illustrate the consistency of functional roles across the network depth.
For each attention map, the $x$-axis represents key positions (all previous frames and the current block) and the $y$-axis represents query positions.
\begin{itemize}
    \item \textbf{Local heads}: exhibit a sharp diagonal pattern with attention concentrated on the current block and its immediate neighbor, with virtually no attention to distant history.
    \item \textbf{Anchor heads}: show a prominent vertical stripe at the first-frame position alongside moderate local attention, confirming the anchoring behavior at multiple network depths.
    \item \textbf{Memory heads}: display a diffuse attention pattern spanning the entire historical context, with attention mass distributed relatively uniformly across previous frames.
\end{itemize}

\subsection{Stability Validation}
\label{sec:supp_stability}

A critical question for the practical applicability of Head Forcing is whether the discovered head classification is stable: does it represent an intrinsic  property of the model, or could it shift under different generation conditions?
We systematically validate stability along three axes: across AR steps, across denoising timesteps, and across text prompts.

We repeat the offline profiling procedure under multiple controlled variations and compare the resulting head partitions. Let
$\mathcal{H}_{\mathrm{local}}^{(r)},
\mathcal{H}_{\mathrm{anchor}}^{(r)},
\mathcal{H}_{\mathrm{mem}}^{(r)}$
denote the three head sets obtained from a profiling run $r$, where a run is defined by a specific choice of prompt subset, AR step subset, denoising step subset.
To quantify stability, we independently classify all 360 heads under each condition and compute defined core stability ratio. Specifically, for each role $c \in \{ \mathrm{local}, \mathrm{anchor}, \mathrm{memory}\}$ and profiling runs (different conditions) $r\in \{1, \cdots, M\}$, we compute the core stability ratio for each head category $c$ as follows:
\[
S_{c}
=
\frac{
\left|\mathcal{H}_{c}^{(1)} \cap \mathcal{H}_{c}^{(2)} \cap \cdots \cap \mathcal{H}_{c}^{(M)} \right|
}{
\frac{1}{M}\sum_{r\in \{1,\cdots,M\}} |\mathcal{H}_{c}^{(r)}|
},
\]
where $\frac{1}{M}\sum_{r\in \{1,\cdots,M\}} |\mathcal{H}_{c}^{(r)}|$ is the average number of head assigned to head role $c$. $S_c$ denotes the proportion of attention heads in category $c$ that consistently appear across different conditions, where a higher $S_c$ indicates more stable profiling results for that head category. We report the core stability ratio per head category to characterize the stability of each individual head category, and define the average core stability ratio as $S_{\text{avg}} = \frac{1}3{}\sum_{c \in \{\text{anchor}, \text{local}, \text{memory}\}} S_c$ to measure the overall stability across all categories.

In our experiments, we set the total number of runs (conditions) to $M = 4$, consistent with the standard practice in existing models that typically perform $4$ denoising steps. This choice ensures that the three condition types that AR steps, denoising steps, and text prompts are evaluated at the same value scale, enabling fair and comparable stability measurements across conditions.

\begin{table}[t]
    \centering
    \caption{\textbf{Core stability ratio $S_c$ across different conditions.}
    }
    \label{tab:core_stability_ar}
    \small
    \resizebox{\textwidth}{!}{%
    \begin{tabular}{lcccc}
    \toprule
    & Anchor Head $S_\text{anchor}$ &  Memory Head $S_\text{memory}$ & Local Head $S_\text{local}$& Average  $S_\text{avg}$ \\
    \midrule
    $S_c$ for different AR step    & $0.901$ & $0.916$ & $0.922$ & $0.913$\\
    $S_c$ for different denoising ste p& $0.949$ & $0.953$& $0.961$ & $0.954$ \\
    $S_c$ for different prompts    & $0.908$& $0.891$ & $0.897$ & $0.898$\\
    \bottomrule
    \end{tabular}}
    \vspace{-10pt}
\end{table}

\noindent\textbf{Stability Across Autoregressive Steps.}
We isolate the profiling statistics at individual autoregressive (AR) steps rather than averaging across them. Specifically, for a fixed prompt and a fixed $3$rd denoising step, we compute the core stability ratio $S_c$ for each head category across four AR steps $i \in \{3, 12, 21, 30\}$. As reported in the first row of Tab.~\ref{tab:core_stability_ar}, $S_c$ remains consistently high (above $0.9$) for all three head categories, indicating that the category assignment of each head is stable across all AR steps. The classification results are nearly identical across steps, with only minor boundary fluctuations observed for a small number of heads near the classification thresholds between memory heads and anchor heads.

\noindent\textbf{Stability Across Denoising Steps.}
Self-Forcing adopts a 4-step denoising schedule ($t \in \{0, 1, 2, 3\}$).
For a fixed AR step and a fixed prompt, we perform head classification independently at each denoising timestep and compute the core stability ratio $S_c$ for each head category.
As reported in the second row of Tab.~\ref{tab:core_stability_ar}, $S_c$ remains consistently high across all denoising timesteps, indicating that the category assignment of each head is equally stable with respect to the denoising stage.
These results confirm that the head classification is robust to variations in the denoising schedule, further supporting the reliability of the proposed profiling method.

\noindent\textbf{Stability Across Prompts.}
Text prompts introduce the most significant source of semantic variation, as different prompts may activate distinct attention patterns. For a fixed AR step and a fixed denoising step, we perform head classification independently on four different text prompts and compute the core stability ratio 
$S_c$ for each head category. As reported in the third row of Tab.~\ref{tab:core_stability_ar}, $S_c$ remains consistently high across all prompt conditions, with only minor fluctuations observed for anchor heads near the anchor–memory decision boundary. These results confirm that the head classification is robust to variations in input semantics, demonstrating that the discovered head partition reflects an intrinsic structural property of the model rather than an artifact of any particular prompt.

\noindent\textbf{Summary of Stability Results.}
Under all three evaluated conditions, namely varying AR steps, denoising timesteps, and text prompts, the core stability ratio $S_c$ remains consistently high for all three head categories, as reported in Tab.~\ref{tab:core_stability_ar}. The average stability ratio $S_{\text{avg}}$ further confirms the overall robustness of the head classification. These results demonstrate that the profiling-based head categorization is stable under diverse inference conditions, and thus head categories can be reliably determined through offline profiling alone, without requiring any additional classification overhead during inference.

\subsection{Computational Overhead of Profiling Process}
\label{sec:computational}
In the standard setting of this work, the full profiling procedure requires generating $20 \times 10 \,\text{s}$ of video while recording attention distributions, and the whole process taking approximately $1.2$ GPU-hours on a single NVIDIA A100-80GB. Once completed, the resulting head classification is stored as a static configuration file and reused across all subsequent inference runs without modification. Importantly, this is a one-time offline cost that is amortized over all subsequent generation runs. Furthermore, we provide precomputed profiling results for popular models, so that users are not required to run the profiling procedure themselves. In practice, owing to the stability of profiling results discussed above, users who choose to run profiling on their own can obtain reliable head classifications with as few as 3 to 4 prompts, reducing the total profiling time to approximately 10 minutes.

\section{Head Profiling on Other AR Video Models}

\noindent\textbf{Setup.}
We apply the same profiling protocol described in Sec.~3.2 of the main paper to LongLive~\cite{yang2025longlive} (an AR video DiT with chunk-wise generation). For each model, we generate 20 diverse prompts and collect per-head attention statistics averaged over AR steps, denoising steps, and prompts.

\begin{figure}[t]
    \centering
    \includegraphics[width=\linewidth]{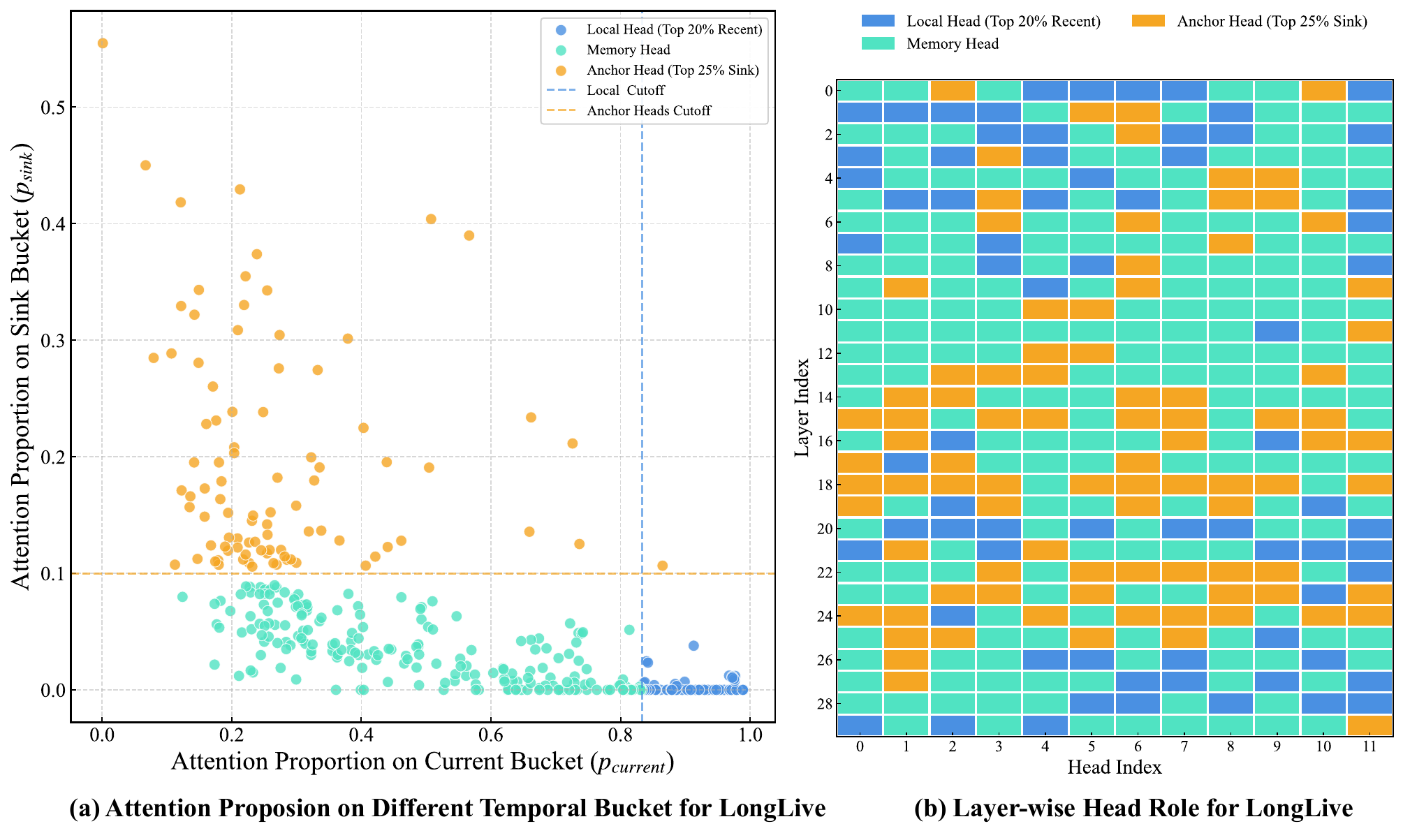}
    \caption{\textbf{Attention head profiling results for LongLive.}}
    \label{fig:head_longlive}
\end{figure}

\noindent\textbf{Profiling Results for LongLive.}
Fig.~\ref{fig:head_longlive} shows the attention proportion scatter plots for all three models. In each case, heads cluster into three clearly separable groups corresponding to local, anchor, and memory heads, confirming that the tripartite head taxonomy is not an artifact of Self Forcing but an intrinsic property shared across AR video DiTs.

\begin{figure}[t]
    \centering
    \includegraphics[width=1.05\linewidth]{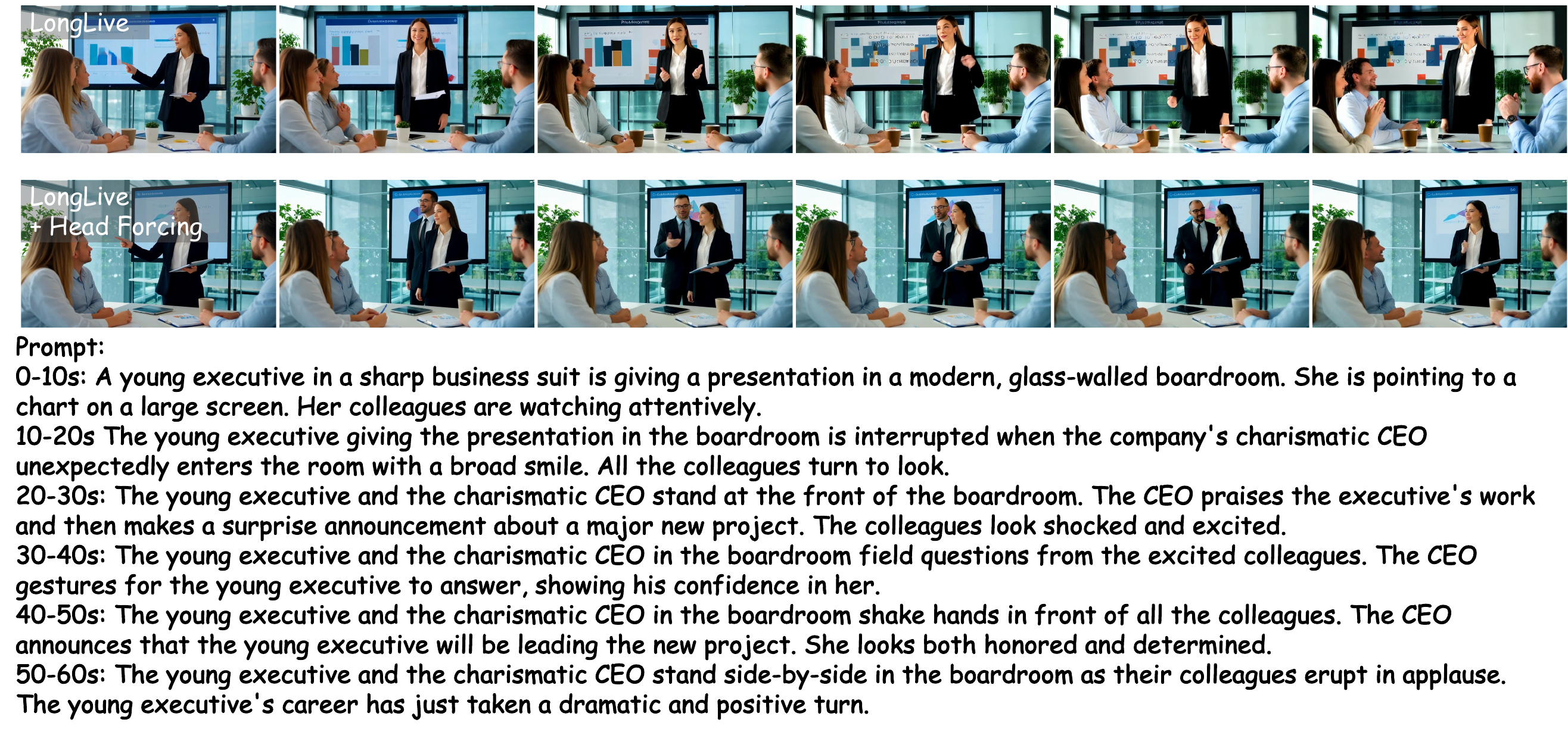}
    \caption{Qualitative results of comparison between LongLive implemented with Head Forcing and vanilla LongLive.}
    \label{fig:headforcing_longlive}
\end{figure}

\noindent\textbf{Head Forcing for LongLive.}
We then use similar Head Forcing method on LongLive with the same setting on Self Forcing.
Fig.~\ref{fig:headforcing_longlive} shows the qualitative results of comparison between LongLive implemented with Head Forcing and vanilla LongLive in multi-prompt guided long video generation. The results show that LongLive implemented with Head Forcing achieves stronger prompt adherence, higher visual quality, and improved temporal consistency.

\section{More Analysis and Result of RoPE Re-encoding}

\noindent\textbf{Comparison with prior RoPE strategies.}
We compare our head-wise RoPE re-encoding against two representative inference-time RoPE strategies.
Deep Forcing~\cite{yi2025deep} stores position-encoded KV cache and applies post-hoc temporal corrections when positions become stale.
Infinity-RoPE~\cite{yesiltepe2025infinity} re-indexes the global sequence uniformly across all heads, resetting sink tokens to position zero.
Both methods treat every head identically, assigning the same temporal indices regardless of the head's functional role or cache content.

Our approach differs in two key aspects.
\textit{First}, we cache keys and values with \emph{spatial RoPE applied}, stripping the temporal component at write time.
This is particularly important for compressed summary tokens in episodic memory: because compression merges spatial tokens from different frames, their spatial semantics should be preserved intact while temporal positions are freely reassignable.
\textit{Second}, re-encoding is performed \emph{per-head} after cache assembly. Since each head type maintains a different cache composition (local heads: $f{+}1$ frames; anchor heads: $2f{+}1$ frames; memory heads: $B_\text{epi}{+}B_\text{fast}{+}f$ frames), the assigned index range and hence the maximum relative position differ across heads. This ensures that \emph{every} head's relative positions stay strictly within the pretrained range, whereas uniform re-indexing can only guarantee this for one cache configuration.

\noindent\textbf{Quantitative impact.}
Table~\ref{tab:rope_ablation} extends the main paper's component ablation (Table~4, row C$\to$D) with a finer breakdown.
RoPE re-encoding yields consistent gains across all metrics, with the largest improvements on \textit{Temporal Flickering} (+0.21) and \textit{Dynamic Degree} (+2.28).
This aligns with our analysis: positional O.O.D.\ primarily manifests as inter-frame inconsistencies (flickering) and unnatural motion dynamics, both of which are mitigated by keeping relative positions in distribution.

\begin{table}[t]
\centering
\caption{Detailed ablation of RoPE re-encoding on 60\,s single-prompt generation (extending Table~4, row C$\to$D).}
\label{tab:rope_ablation}
\resizebox{\linewidth}{!}{
\begin{tabular}{lcccccc}
\toprule
 & Dynamic & Motion & Temporal & Subject & Imaging & Aesthetic \\
 & Degree$\uparrow$ & Smoothness$\uparrow$ & Flickering$\uparrow$ & Consistency$\uparrow$ & Quality$\uparrow$ & Quality$\uparrow$ \\
\midrule
w/o RoPE re-enc. & 39.09 & 98.54 & 98.51 & 97.73 & 68.93 & 61.04 \\
w/ RoPE re-enc. (Ours) & \textbf{41.37} & \textbf{98.69} & \textbf{98.72} & \textbf{97.81} & \textbf{69.27} & \textbf{61.36} \\
\bottomrule
\end{tabular}}
\end{table}

\begin{figure}[t]
    \centering
    \includegraphics[width=0.9\linewidth]{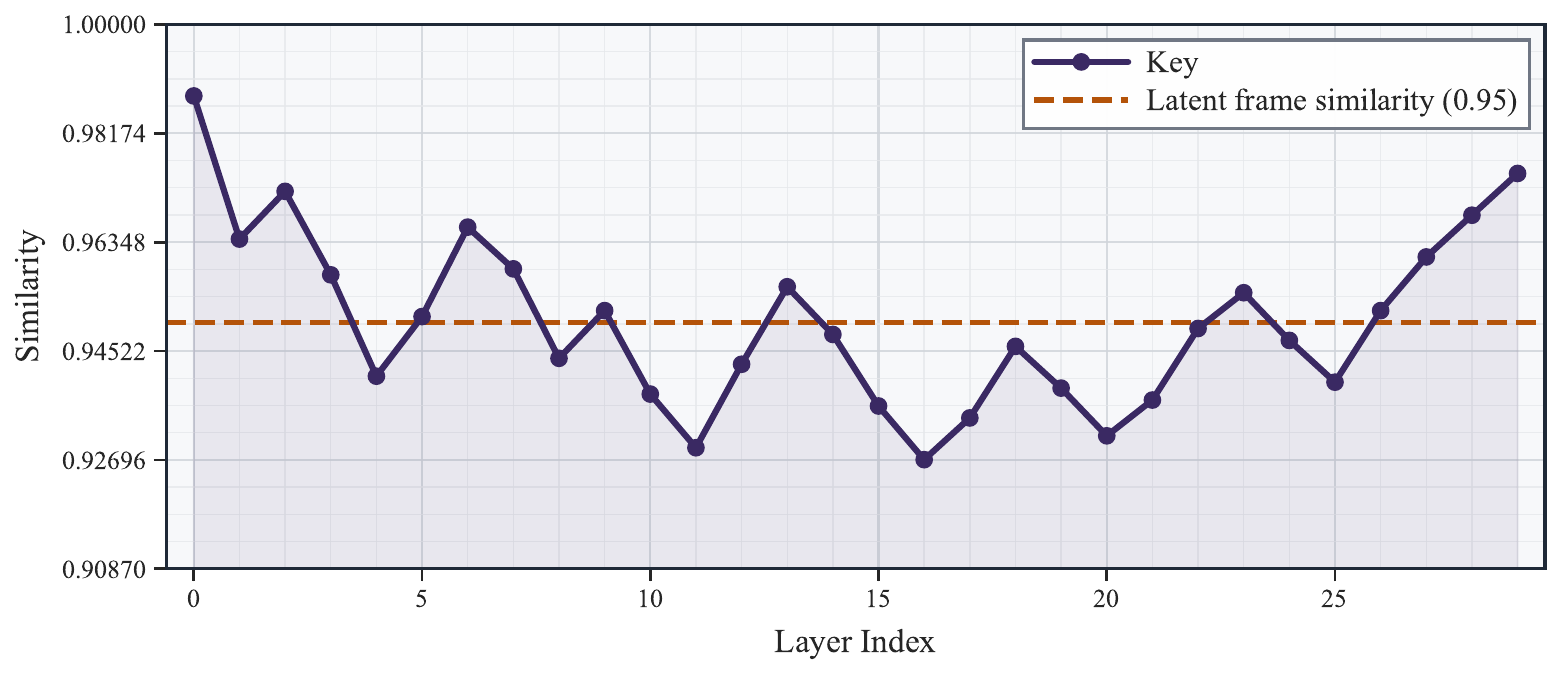}
    \caption{When the cosine similarity between two latent frames is approximately 0.95, we compute the key embedding similarity across all layers for the corresponding frame pair.}
    \label{fig:key_embedding}
\end{figure}

\section{Effect of Key Embeddings}
\noindent\textbf{Why key embeddings?}
In causal DiTs, keys encode what information a token offers to be attended to, making them natural semantic descriptors of frame content.
Moreover, key embeddings are readily available in the KV cache without additional computation, whereas alternatives such as latent frame similarity would require extra storage of the corresponding latent frames and computation.
Prior work on KV cache compression for video understanding~\cite{kim2025infinipot} has also demonstrated that key-based similarity reliably captures temporal redundancy across frames, motivating our adoption.

As shown in Fig.~\ref{fig:key_embedding}, when the cosine similarity between two latent frames is approximately 0.95, we compute the key embedding similarity across all layers for the corresponding frame pair. The resulting key embedding similarity closely mirrors the latent frame similarity, indicating that key embedding similarity can serve as a reliable proxy for latent frame similarity in measuring inter-frame resemblance.

\noindent\textbf{Alternative metrics.}
We compare these two similarity metrics for computing the novelty score $\delta(K_c)$ in Eq.~4, keeping all other components (head-wise allocation, hierarchical memory, RoPE re-encoding) and hyperparameters ($\tau_\text{novel}=0.95$) fixed:
\begin{enumerate}
    \item \textbf{Key cosine} (Ours): cosine similarity on mean-pooled key embeddings across memory heads and layers, as described in Eq.~4.

    \item \textbf{Latent frame similarity}: computing candidate frame cosine similarity between candidate frame and latent episodic memory frames.
\end{enumerate}

\noindent\textbf{Results.}
Table~\ref{tab:novelty_metric} reports 60\,s generation results under both single-prompt and multi-prompt settings. 
Key cosine similarity performs on par with the substantially more expensive  latent frame similarity across all indicators, confirming that key embeddings are sufficiently expressive proxies for frame-level semantics. 

\begin{table}[h]
\centering
\caption{Ablation on scene novelty metric for episodic memory admission.} 
\label{tab:novelty_metric}
\resizebox{\linewidth}{!}{%
\begin{tabular}{l cc cc cc c}
\toprule
 & \multicolumn{4}{c}{\textbf{60\,s Single-Prompt}} & \multicolumn{3}{c}{\textbf{60\,s Multi-Prompt}} \\
\cmidrule(lr){2-5} \cmidrule(lr){6-8}
\textbf{Metric} & Dynamic & Motion & Subject & Imaging & Quality & Consistency & CLIP Score \\
 & Degree$\uparrow$ & Smooth.$\uparrow$ & Consist.$\uparrow$ & Quality$\uparrow$ & Score$\uparrow$ & Score$\uparrow$ & Average$\uparrow$ \\
\midrule
CLIP similarity & 41.42 & 98.62  & 97.80 & 69.33 & 84.68 &  96.73 & 24.98  \\
Key cosine (Ours) & 41.37 & 98.69 & 97.81 & 69.27& 84.74 & 96.69 & 24.95 \\

\bottomrule
\end{tabular}%
}
\vspace{-2mm}
\end{table}
This confirms that key embeddings strike an effective balance between semantic expressiveness and computational efficiency, making them well-suited for online episodic memory management in real-time generation.

\section{Extended Scalability Experiments}

\begin{figure}[h]
\centering
\includegraphics[width=\linewidth]{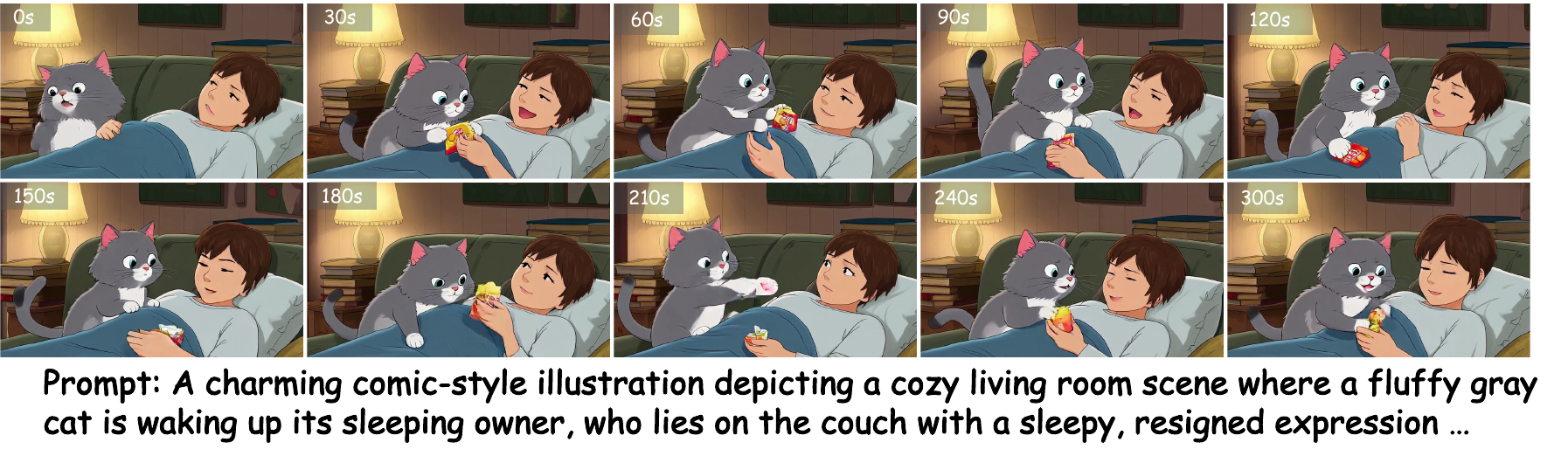}
\caption{\textbf{Qualitative results on 5min ultra long video generation.}
Head Forcing maintains visual fidelity throughout the video sequence.}
\label{fig:qual_extended}
\end{figure}

\noindent\textbf{Qualitative Results on Extended Generation (Ultra Long Video Generation).}
Head Forcing is able to scale to longer horizons through head-wise KV cache allocation  and dynamic episodic memory management.
Fig.~\ref{fig:qual_extended} presents sampled keyframes from 5min (300\,s) single-prompt ultra long video generation.
Head Forcing is able to preserve subject identity and visual quality throughout the long rollout.

\begin{figure}[h]
\centering
\includegraphics[width=0.8\linewidth]{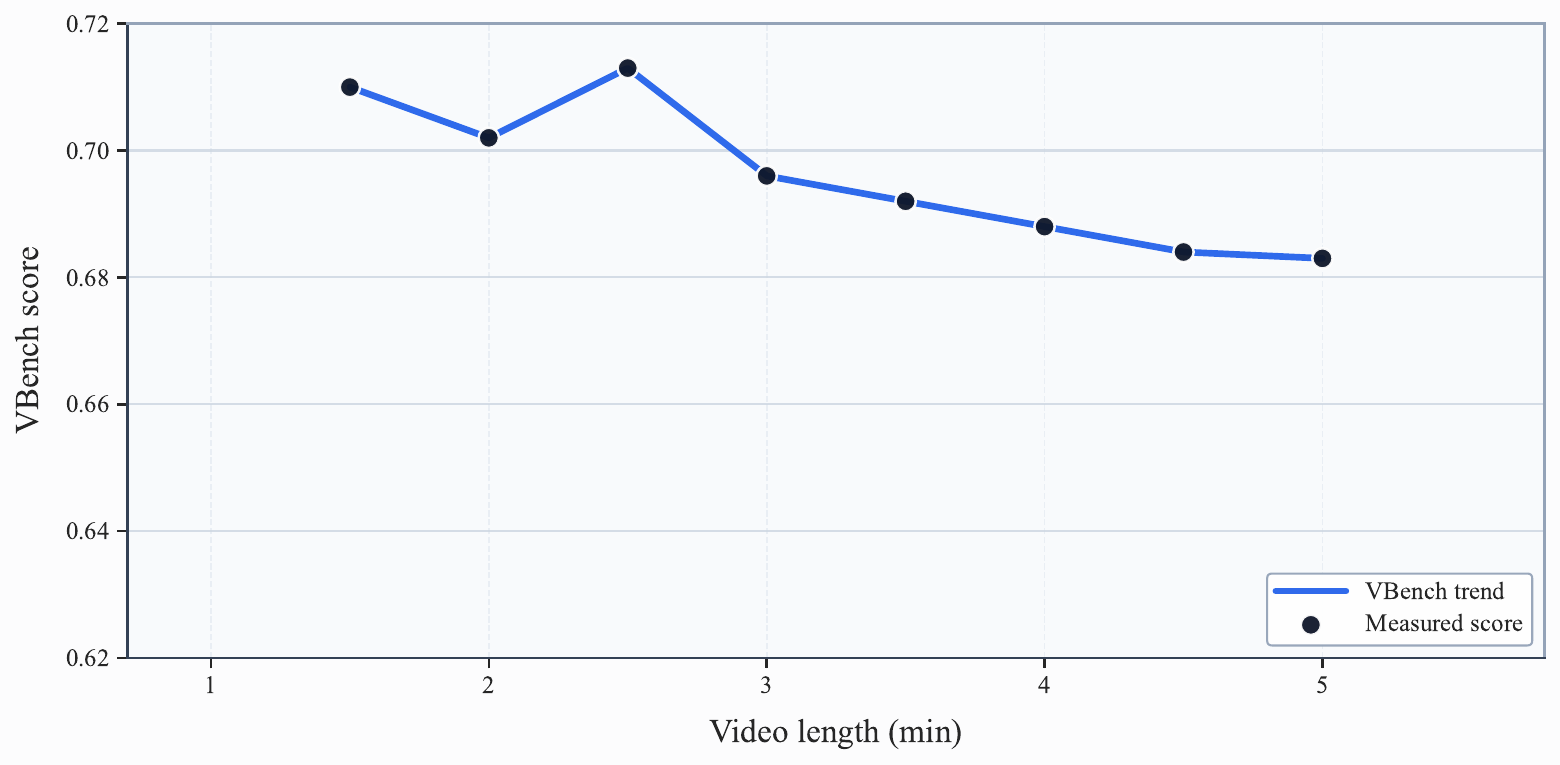}
\caption{\textbf{Average VBench-Long score across different video lengths..}
}
\label{fig:curve_long}
\end{figure}

\noindent\textbf{Quantitative Results on Ultra-Long Video Generation.}
Fig.~\ref{fig:curve_long} presents the average VBench-Long score~\cite{huang2023vbench} as a function of video length under the single-prompt long video generation setting.  The scores do not exhibit a sharp decline as the video length grows, remaining stable at a consistently high level. This demonstrates that Head Forcing can preserve overall video quality even under substantially extended generation lengths.

\section{Additional Ablation Studies}

\subsection{Qualitative Results from Ablation Study on Main Components}
\begin{figure}[h]
\centering
\includegraphics[width=1.02\linewidth]{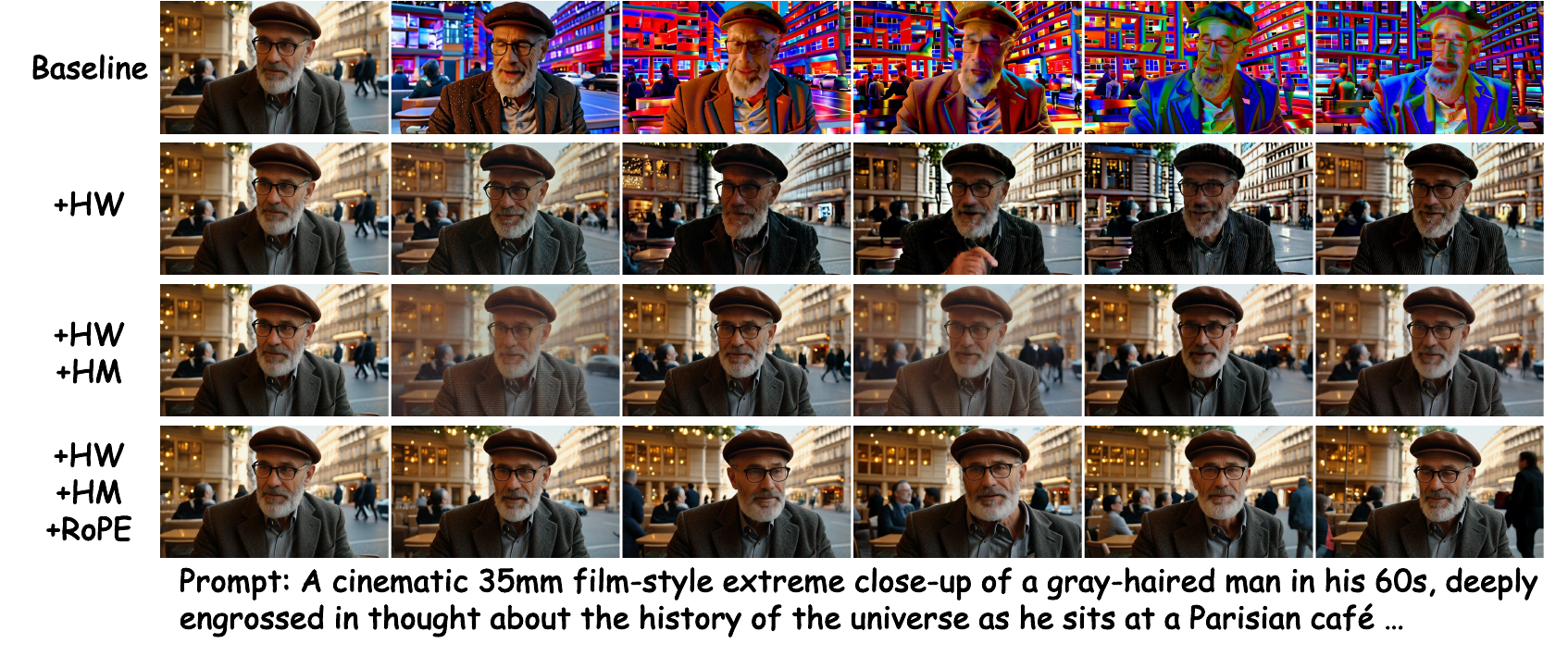}
\caption{\textbf{Qualitative results from the ablation study of main components.}
HW=head-wise kv
cache allocation, HM = hierarchical memory, RoPE = head-wise RoPE re-encoding.
}
\label{fig:vis_ablation}
\end{figure}

In Fig.~\ref{fig:vis_ablation}, we present additional qualitative results from the ablation study of main components on 60\,s long video generation. Incorporating head-wise allocation improves visual quality, consistency, and motion dynamics. Adding the hierarchical memory system further enhances contextual consistency across the generated sequence. Finally, introducing head-wise re-encoding reduces flickering artifacts.

\subsection{Additional Ablation Studies on Hyperparameters}

We provide comprehensive sensitivity analysis for other hyperparameters, extending the analysis of $\tau_{\text{local}}$ and $B_{\text{epi}}$ in main paper Fig.~6.

\begin{figure}[t]
    \centering
    \includegraphics[width=0.9\linewidth]{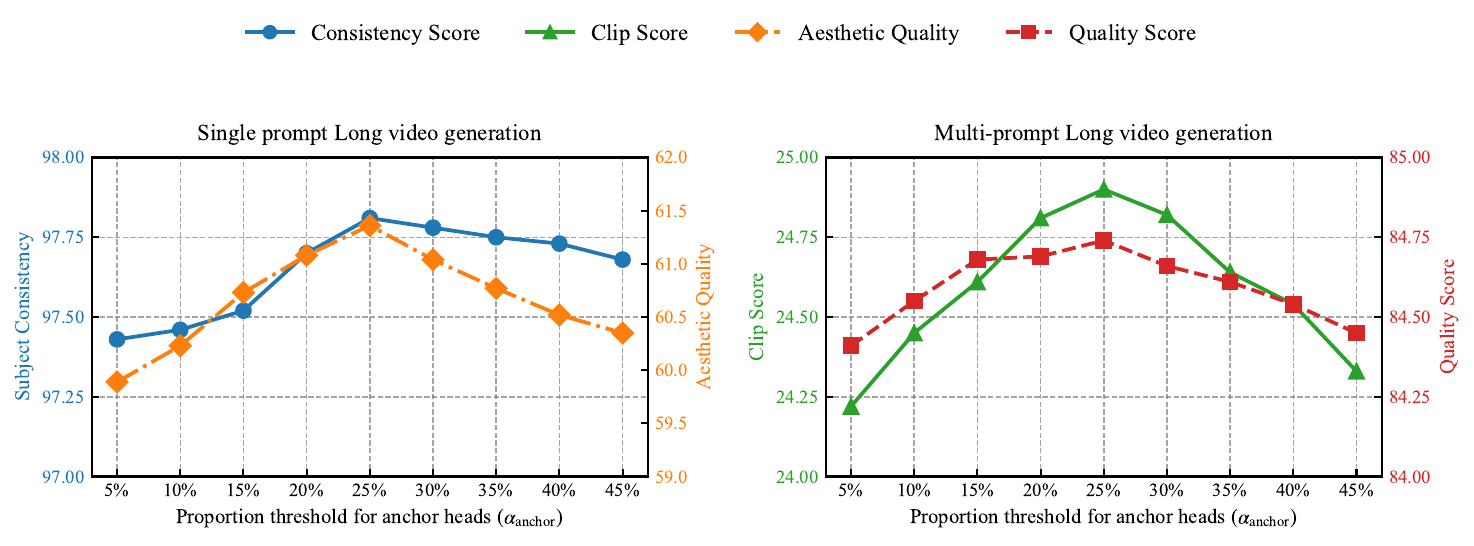}
    \caption{\textbf{Impact of $\alpha_{\text{anchor}}$ on 60\,s generation quality.}
    Performance is robust within $[20\%, 35\%]$.
    Imaging quality (strongly correlated with visual stability) is most sensitive, consistent with anchor heads' role in preventing color drift.
    }
    \label{fig:supp_sensitivity_anchor}
    \vspace{-10pt}
\end{figure}

\noindent\textbf{Sensitivity of $\alpha_{\text{anchor}}$ (Anchor Head Threshold).}
This analysis was deferred from the main paper and is essential for validating the robustness of anchor head selection.
We vary $\alpha_{\text{anchor}} \in \{5\%, 10\%, 15\%,\cdots, 40\%, 45\%\}$, keeping other parameter fixed.
Fig.~\ref{fig:supp_sensitivity_anchor} presents the results.
Performance is robust across $\alpha_{\text{anchor}} \in [20\%, 35\%]$.
Setting $\alpha_{\text{anchor}}$ too low ($<$10\%) fails to identify enough anchor heads, losing the stabilization benefit.
Setting it too high ($>$40\%) over-assigns anchor heads, reducing the number of memory heads and thus the capacity for long-range context aggregation.
In particular, the imaging quality metric is most sensitive to $\alpha_{\text{anchor}}$, as insufficient anchoring leads to color drift, while excessive anchoring reduces the expressiveness of the memory system.

\begin{figure}[t]
    \centering
    \includegraphics[width=0.9\linewidth]{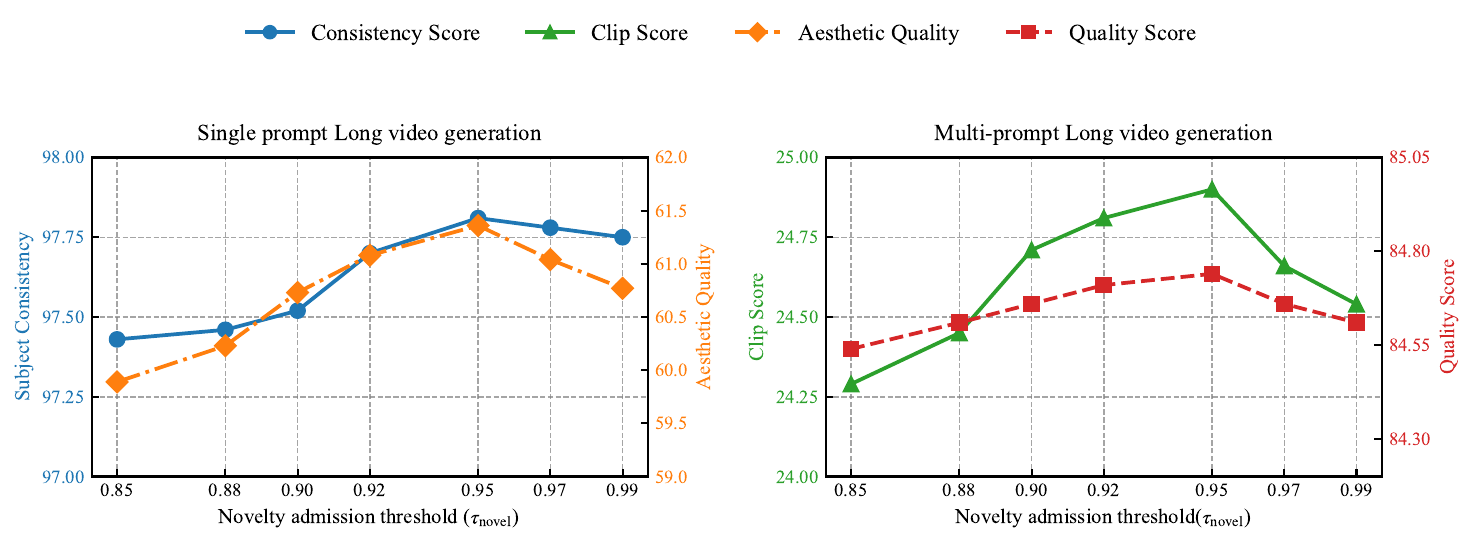}
    \caption{\textbf{Sensitivity of $\tau_{\text{novel}}$.}
    }
    \label{fig:supp_sensitivity_novel}
    \vspace{-10pt}
\end{figure}

\noindent\textbf{Sensitivity of $\tau_{\text{novel}}$ (Novelty Admission Threshold)}
We vary $\tau_{\text{novel}} \in \{0.80, 0.85, 0.88, 0.90, 0.92, 0.95, 0.97, 0.99\}$, keeping all other hyperparameters fixed.
We evaluate on both single-prompt 60\,s and multi-prompt 60\,s settings.
Fig.~\ref{fig:supp_sensitivity_novel} presents the results.
Performance is stable in the range $\tau_{\text{novel}} \in [0.90, 0.97]$.
A threshold that is too permissive ($\tau_{\text{novel}} \ge 0.99$) admits too few frames into episodic memory, effectively disabling it.
A threshold that is too aggressive ($\tau_{\text{novel}} \le 0.85$) admits too many frames, causing frequent compression and diluting the quality of stored entries.
Notably, the multi-prompt CLIP score is more sensitive than single-prompt metrics, as the episodic memory is critical for maintaining cross-prompt contextual awareness.

\begin{figure}[t]
    \centering
    \includegraphics[width=0.9\linewidth]{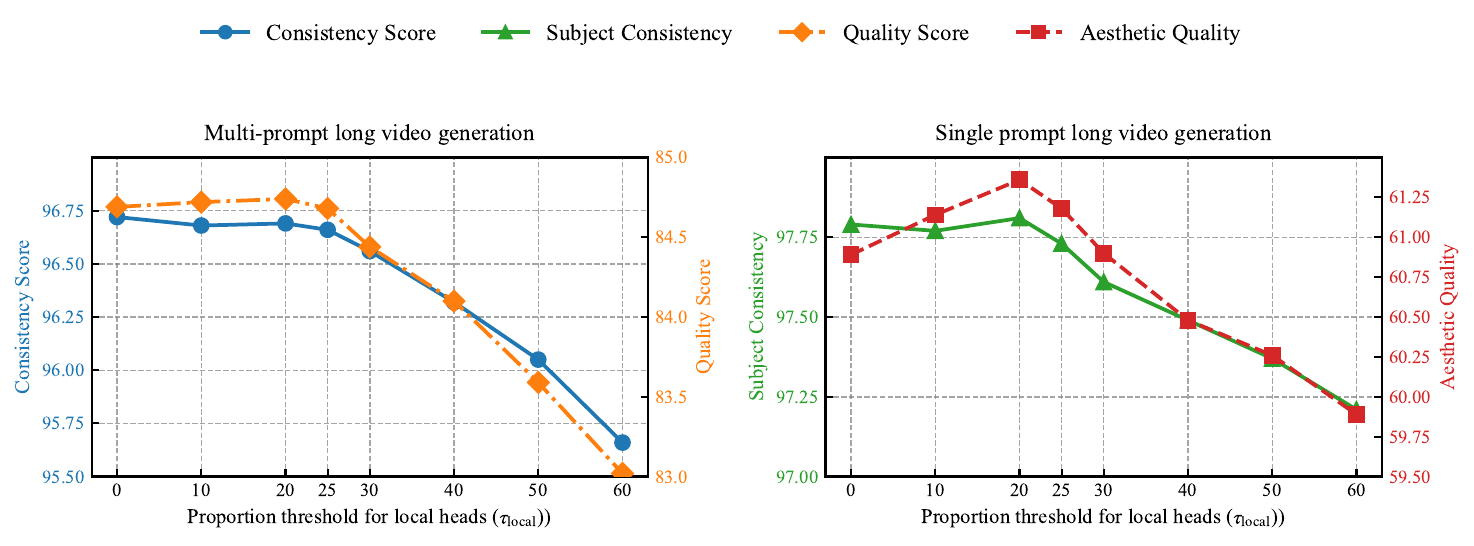}
    \caption{\textbf{Sensitivity of $\tau_{\text{local}}$.}}
    \label{fig:supp_sensitivity_local}
    \vspace{-15pt}
\end{figure}

\noindent\textbf{More Results for Sensitivity of $\tau_{\text{local}}$}:
In the main paper, we presented a sensitivity analysis of $\tau_{\text{local}}$, focusing on the multi-prompt setting. Here, we additionally provide
results for single-prompt long video generation in Fig.~\ref{fig:supp_sensitivity_local} combined with multi-prompt interactive video generation.

\subsection{Impact of Component of Hierarchical Memory System}
\begin{figure}[htp!]
    \centering
    \includegraphics[width=1\linewidth]{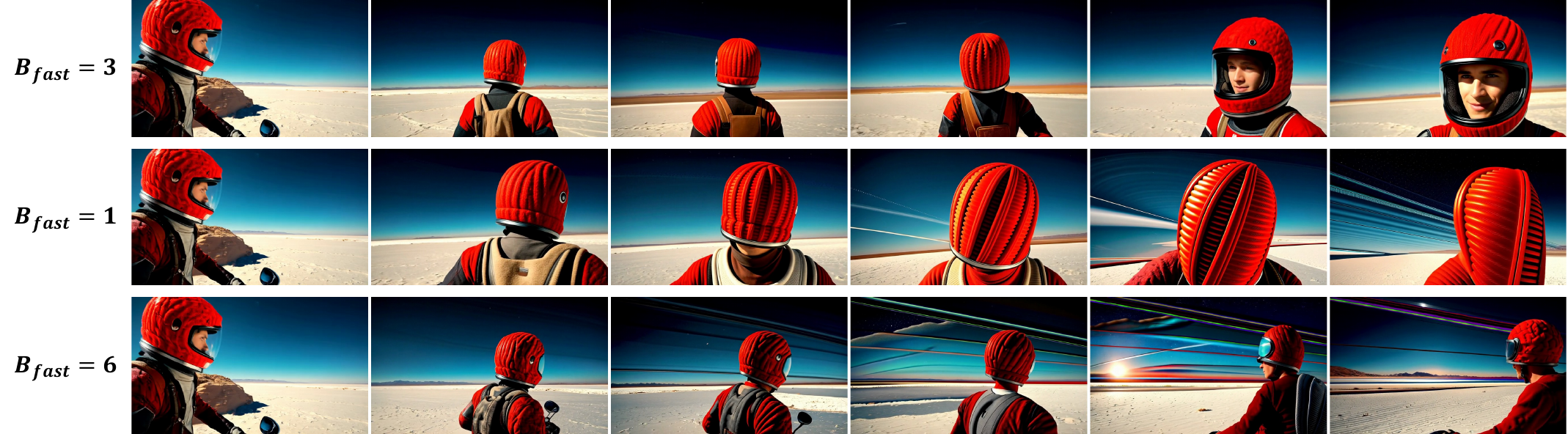}
    \caption{\textbf{Impact of number of fast memory $B_\text{epi}$}.}
    \label{fig:fast_mem}
    \vspace{-15pt}
\end{figure}

\noindent\textbf{Fast Memory.}
The fast memory $\mathcal{M}_\text{fast}$ provides memory heads with fine-grained short-term context for motion continuity and visual coherence. We vary $B_\text{fast}$ (the number of retained recent frames) from 0 to 6, as shown in Fig.~\ref{fig:fast_mem}.
When $B_\text{fast}=1$, memory heads have minimal access to the immediate temporal neighborhood beyond the current block, which markedly impairs motion smoothness and introduces visible artifacts. A moderate value of $B_\text{fast}=3$ is sufficient to maintain smooth temporal transitions. Further increasing $B_\text{fast}$ to $6$ yields diminishing returns and eventually degrades performance, as the growing proportion of fast memory tokens amplifies noise propagation.

\begin{figure}[htp!]
    \centering
    \includegraphics[width=1\linewidth]{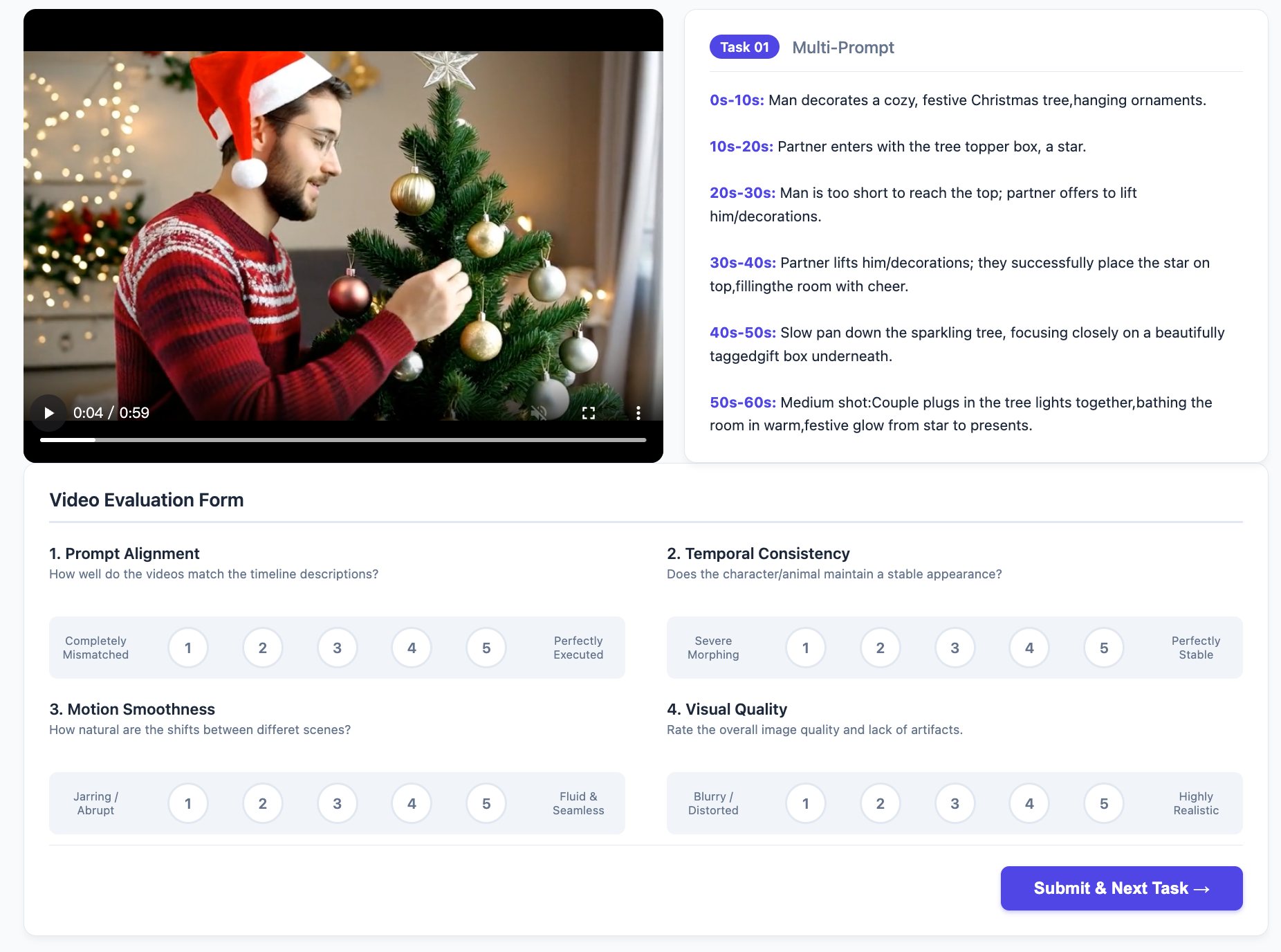}
    \caption{\textbf{User study interfaces}}
    \label{fig:user_study}
    \vspace{-15pt}
\end{figure}

\section{User Study}

In Fig.~\ref{fig:user_study}, we present one of the example interfaces used in our user studies (for multi-prompt interactive video generation).

\noindent\textbf{Single-Prompt Long Video Generation.}
We conduct a user study on 60\,s single-prompt video generation, comparing Head Forcing against Self Forcing~\cite{huang2025self}, LongLive~\cite{yang2025longlive}, Rolling Forcing~\cite{liu2025rolling}, Deep Forcing~\cite{yi2025deep},  and Infinity-RoPE~\cite{yesiltepe2025infinity}. We randomly select 15 prompts from MovieGenBench~\cite{polyak2024movie} and generate one video per method per prompt. We recruit 20 participants, each rating every video independently on a 5-point Likert scale across three aspects: (i)~Visual Quality: overall sharpness and absence of artifacts, (ii)~Temporal Consistency: identity preservation and absence of flickering or drift, and (iii)~Motion Dynamics: richness and naturalness of motion. Results are reported in Tab.~\ref{tab:user_single}. Head Forcing achieves the highest mean score across all three criteria, with particularly strong advantages in Motion Dynamics and Temporal Consistency, consistent with our automatic evaluation results.

\begin{table}[h]
\centering
\vspace{-10pt}
\caption{\textbf{User study on single-prompt 60\,s generation.} 5-point Likert scale; higher is better.}
\label{tab:user_single}
\small
\resizebox{\textwidth}{!}{%
\begin{tabular}{l|cccc}
\toprule
Method & Visual Quality\,$\uparrow$ & Temporal Consistency$\uparrow$ & Motion Smoothness,$\uparrow$& Avg \,$\uparrow$ \\
\midrule
Self Forcing~\cite{huang2025self} & 1.56& 2.09 & 1.92& 1.86\\
 Rolling Forcing~\cite{liu2025rolling} & \textbf{4.12}& 3.71& 2.34&3.39\\
Deep Forcing~\cite{yi2025deep} & 3.74& 3.07& 3.05 & 3.29\\
Infinity-RoPE~\cite{yesiltepe2025infinity} & 3.66& 3.18 & 2.91 & 3.25\\
LongLive~\cite{yang2025longlive} & 3.83& 3.35& 3.21& 3.46\\
\textbf{Head Forcing (Ours)} & 4.04& \textbf{3.98}& \textbf{3.67} & \textbf{3.89}\\
\bottomrule
\end{tabular}}
\end{table}

\noindent\textbf{Multi-Prompt Interactive Video Generation.}
We further evaluate on 60\,s multi-prompt interactive generation using the same participant pool. We select 15 six-prompt sequences from multi-prompt benchmark, comparing Head Forcing against Self Forcing~\cite{huang2025self}, LongLive~\cite{yang2025longlive}, and Infinity-RoPE~\cite{yesiltepe2025infinity}. Participants rate each video on the same three criteria above with an additional dimension: (iv)Prompt Adherence how faithfully each segment reflects its corresponding prompt. As shown in Tab.~\ref{tab:user_multi}, Head Forcing obtains the best scores across all four criteria. The advantage is most pronounced on Prompt Alignment, confirming that our episodic memory system effectively preserves context across prompt transitions.

\begin{table}[h]
\centering
\vspace{-10pt}
\caption{\textbf{User study on multi-prompt 60\,s generation.} 5-point Likert scale; higher is better.}
\label{tab:user_multi}
\resizebox{\textwidth}{!}{%
\small
\begin{tabular}{l|ccccc}
\toprule
Method & Visual Quality\,$\uparrow$ & Temporal\ Consistency\,$\uparrow$ & Motion Smoothness.\,$\uparrow$ & Prompt Alignment\,$\uparrow$ & Avg\,$\uparrow$ \\
\midrule
Self Forcing~\cite{huang2025self} & 1.81& 1.95 & 2.18& 2.20& 2.03\\
Infinity-RoPE~\cite{yesiltepe2025infinity} & 3.15 & 3.21& 2.95& 3.19& 3.12\\
LongLive~\cite{yang2025longlive} & 3.78&  3.08& 3.09& 3.28& 3.30\\
\textbf{Head Forcing (Ours)} & \textbf{3.93}& \textbf{3.39}& \textbf{3.45}& \textbf{3.47} & \textbf{3.56}\\
\bottomrule
\end{tabular}}
\vspace{-15pt}
\end{table}

\section{Efficiency Optimization Validation}
\label{sec:supp_efficiency}

In Section~3.5 of the main paper, we introduce two complementary efficiency optimizations: variable-length FlashAttention and fused Triton kernels to address the computational challenges arising from head-wise KV caches allocation. In this section, we provide a  validation of their effectiveness with throughput analysis across generation lengths. We also compare our method's memory usage with baselines.

\subsection{Throughput With Different Efficiency Optimization}

\noindent\textbf{Implementation.} To isolate the contribution of each optimization, we compare four implementation variants under \emph{identical} head-wise KV cache allocation strategies:
\textbf{(V1) Naive Loop}: Each attention head is computed by a separate flash attention~\cite{dao2022flashattention, dao2023flashattention} call inside a Python \texttt{for}-loop. KV cache updates, sequence assembly, and RoPE re-encoding are likewise performed in a per-head loop using standard PyTorch tensor operations.
\textbf{(V2) + VarLen FlashAttn}: All heads are packed into a single flat buffer with per-head sequence-length metadata and processed via a single \texttt{flash\_attn\_varlen\_func} call, eliminating per-head kernel launches.
\textbf{(V3) + Fused Cache Kernels}: Building on (V2), a Triton kernel is introduced for KV cache rolling updates. For episodic memory updates, we gather all memory heads per layer and pack them into a single tensor, enabling batched computation of novelty scores and prompt-guided compression along the batch dimension. The resulting updates are then applied through a custom Triton kernel for episodic memory management.
\textbf{(V4) Full (Ours)}: Building on (V3), the additional three-stage pipeline containing sequence assembly, RoPE re-encoding, and flat-buffer packing for each head is fused into a single Triton kernel, yielding the complete Head Forcing system.

\noindent\textbf{Throughput.}

\begin{figure}[htp!]
    \centering
    \includegraphics[width=0.95\linewidth]{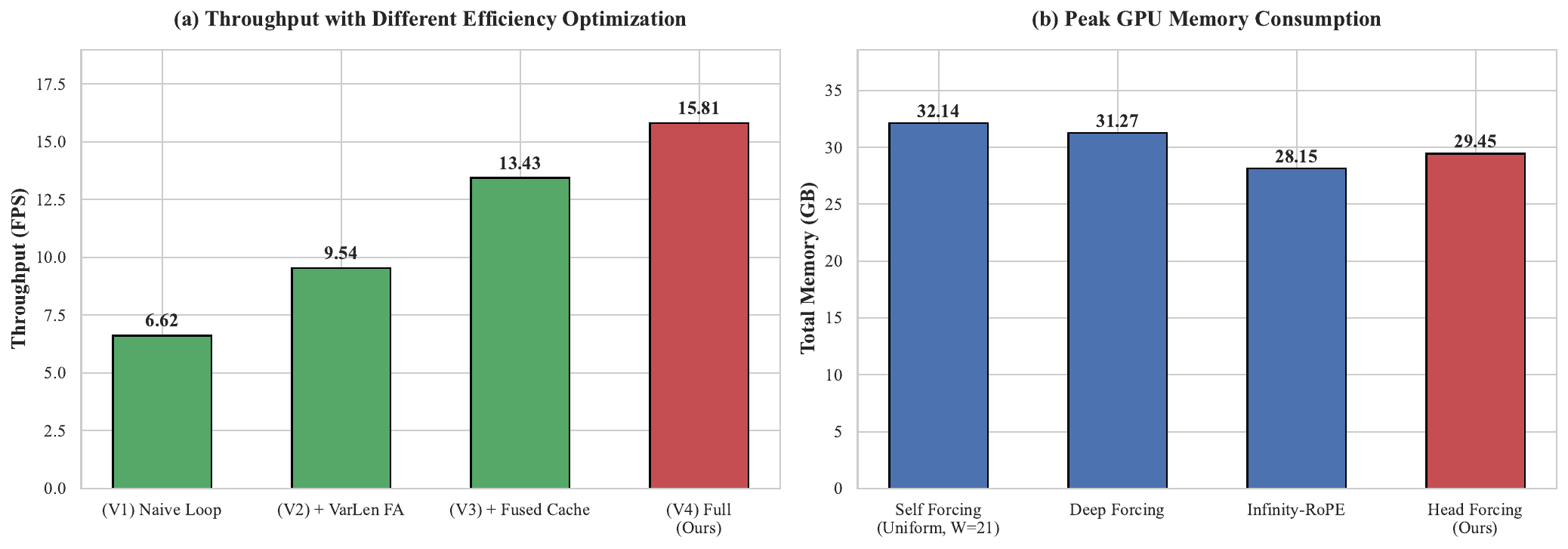}
    \caption{\textbf{Efficiency Optimization Validation.} (a)~Throughput (FPS) with different efficiency optimization. (b)~\textbf{Peak GPU memory consumption} during 60\,s video generation at $832 \times 480$ resolution on a single A100-80GB.}
    \label{fig:efficiency}
    \vspace{-15pt}
\end{figure}

As shown in Fig.~\ref{fig:efficiency} (a), the naive implementation (V1) achieves only 6.62\,FPS, making it unsuitable for real-time applications. Each subsequent optimization stage progressively recovers throughput to 15.81\,FPS, demonstrating that Variable-Length FlashAttention and the fused Triton kernels are essential for making head-wise KV cache management practically efficient.

\subsection{GPU Memory Consumption}
Head-wise cache allocation not only improves generation quality but also offers notable memory efficiency benefits, as local and anchor heads maintain substantially smaller caches than the uniform baseline.
As shown in Fig.~\ref{fig:efficiency}, Head Forcing consumes less GPU memory than the uniform-cache approaches of Self Forcing~\cite{huang2025self} and Deep Forcing~\cite{yi2025deep}, while requiring only slightly more memory than Infinity-RoPE~\cite{yesiltepe2025infinity}. This advantage is particularly beneficial when deploying on memory-constrained GPUs or scaling to higher resolutions.

\section{Memory Budget Fairness Analysis}
\label{sec:supp_budget}

A natural concern is whether the performance gains of Head Forcing stem from a smarter cache allocation strategy or simply from using more total memory than the baselines. 
We address this with a precise per-method budget accounting that shows Head Forcing actually uses \emph{substantially less} total cache.

We express the KV cache budget as the total number of \emph{frame-slots}, defined as $\sum_{h=1}^{N_\text{total}} C_h$, where $C_h$ is the number of cached frames for head $h$ and $N_\text{total} = L \times H = 30 \times 12 = 360$ is the total number of attention heads in the model.
This metric is hardware-agnostic and directly proportional to actual memory consumption, since every frame-slot stores the same number of KV tokens.

Specifically, Self Forcing~\cite{huang2025self} maintains a uniform sliding-window cache of $W = 21$ frames across all heads, with total frames $360 \times 21 = 7{,}560$
Deep Forcing~\cite{yi2025deep}, Infinity-RoPE~\cite{yesiltepe2025infinity} and LongLive~\cite{yang2025longlive} adopt attention-sink strategies that replace a small portion of the sliding window with dedicated sink tokens from the first few frames. Their total per-head cache size is about $16$ for Deep Forcing, $8$ for Infinity-Rope and $12$ for LongLive.

Under the profiling results reported in Sec.~3.2 of the main paper, the 360 heads are categorized as follows (exact counts may vary slightly with profiling prompt set; see Sec.~\ref{sec:supp_stability} for stability analysis):
\begin{itemize}
  \item \textbf{Local heads} ($N_\text{local}$): top $\tau_\text{local}\!=\!20\%$ by current-bucket attention $\;\Rightarrow\;$ \textbf{72 heads}, each caching $f+1 = 4$ frames.
  \item \textbf{Anchor heads} ($N_\text{anchor}$): top $\alpha_\text{anchor}\!=\!25\%$ by first-frame attention $\;\Rightarrow\;$ \textbf{90 heads}, each caching $2f+1 = 7$ frames.
  \item \textbf{Memory heads} ($N_\text{mem}$): remaining $\;\Rightarrow\;$ \textbf{198 heads}, each caching $B_\text{epi}+B_\text{fast}+f = 5+3+3 = 11$ frames.
\end{itemize}
The total budget of Head Forcing is therefore about $\mathbf{3{,}096}\;\text{frame-slots}$.
Tab.~\ref{tab:budget_comparison} summarizes the budget of each method.
Head Forcing uses smaller budget of the total KV cache budget consumed by Self Forcing, Deep Forcing and LongLive as well as comparable budget consumed by Infinity-Rope , while achieving superior generation quality as shown in Tab.~2 of the main paper.
This immediately rules out the hypothesis that the gains simply come from larger memory. On the contrary, Head Forcing achieves better results with the help of better KV cache allocation.

\begin{table}[h]
\centering
\caption{\textbf{KV cache budget comparison.}
All methods are built on the same Self Forcing backbone.
Frame-slots $= \sum_h C_h$ is the total number of cached frames summed across all 360 attention heads. Relative budget is normalized to Head Forcing.}
\label{tab:budget_comparison}
\vspace{2pt}
\small
\begin{tabular}{lccccc}
\toprule
\textbf{Method} & \textbf{Cache per Head} & \textbf{Frame-slots} & \textbf{Relative Budget} \\
\midrule
Self Forcing~\cite{huang2025self}   & 21 (uniform)     & 7,560 & 244.2\% \\
Deep Forcing~\cite{yi2025deep}      & 16 (uniform)     & 5.760 & 186.0\% \\
Infinity-RoPE~\cite{yesiltepe2025infinity} & 8 (uniform)  & 2,880 & 93.0\% \\
LongLive~\cite{yang2025longlive} & 12 (uniform)  & 4,320 & 139.5\% \\
\midrule
\textbf{Head Forcing (Ours)} & 4 / 7 / 11 (head-wise) & \textbf{3,096} & \textbf{100.0\%} \\
\bottomrule
\end{tabular}
\end{table}

\section{Limitations}
\label{sec:limitations}
While Head Forcing achieves strong results on long autoregressive video generation without additional training, several limitations remain.
\textit{(1)~Offline profiling dependency.}
Our method requires a one-time offline profiling step to classify attention heads into local, anchor, and memory types. Although this profiling is lightweight and yields highly stable classifications across different conditions (see Sec.~\ref{sec:supp_stability}), it must be re-executed for each new base model. Developing a universal, architecture-agnostic head classification rule or profiling-free methods remains an open direction.

\textit{(2)~Episodic memory information loss under extreme length.}
The prompt-guided compression mechanism enables episodic memory to operate within a fixed budget indefinitely. However, after many compression cycles over very long sequences (\eg, $>$5 minutes), the summary frame may gradually lose fine-grained details of early scenes. This is an inherent trade-off between memory capacity and generation length that applies to all bounded-memory approaches.

\section{Additional Visual Results}
\begin{figure}[!htbp]
    \centering
    \includegraphics[width=1.05\linewidth]{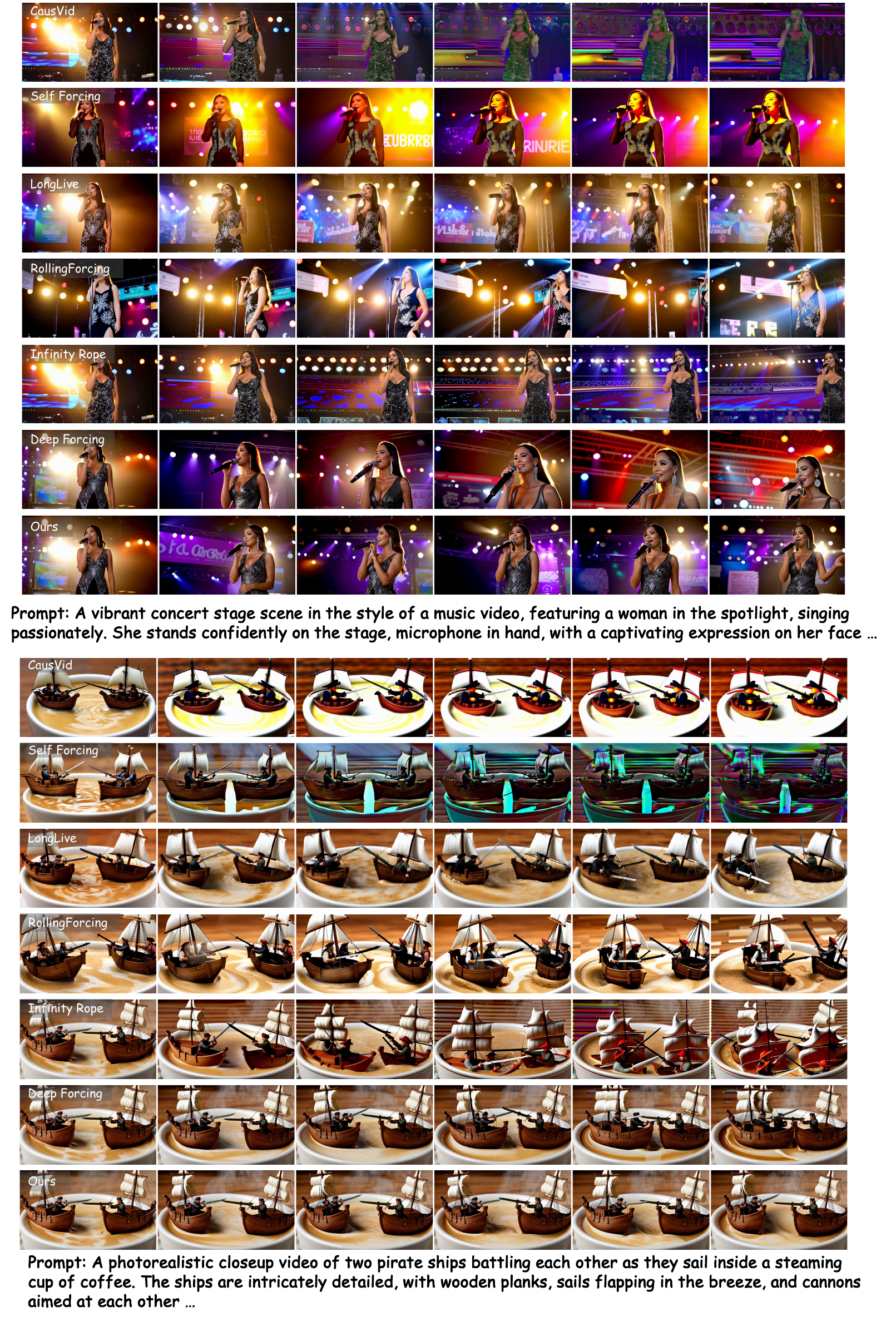}
    \caption{\textbf{Qualitative results on single-prompt long video generation (60\,s)}}
    \label{fig:quality_1}
    \vspace{-15pt}
\end{figure}

\begin{figure}[!htbp]
    \centering
    \includegraphics[width=1.05\linewidth]{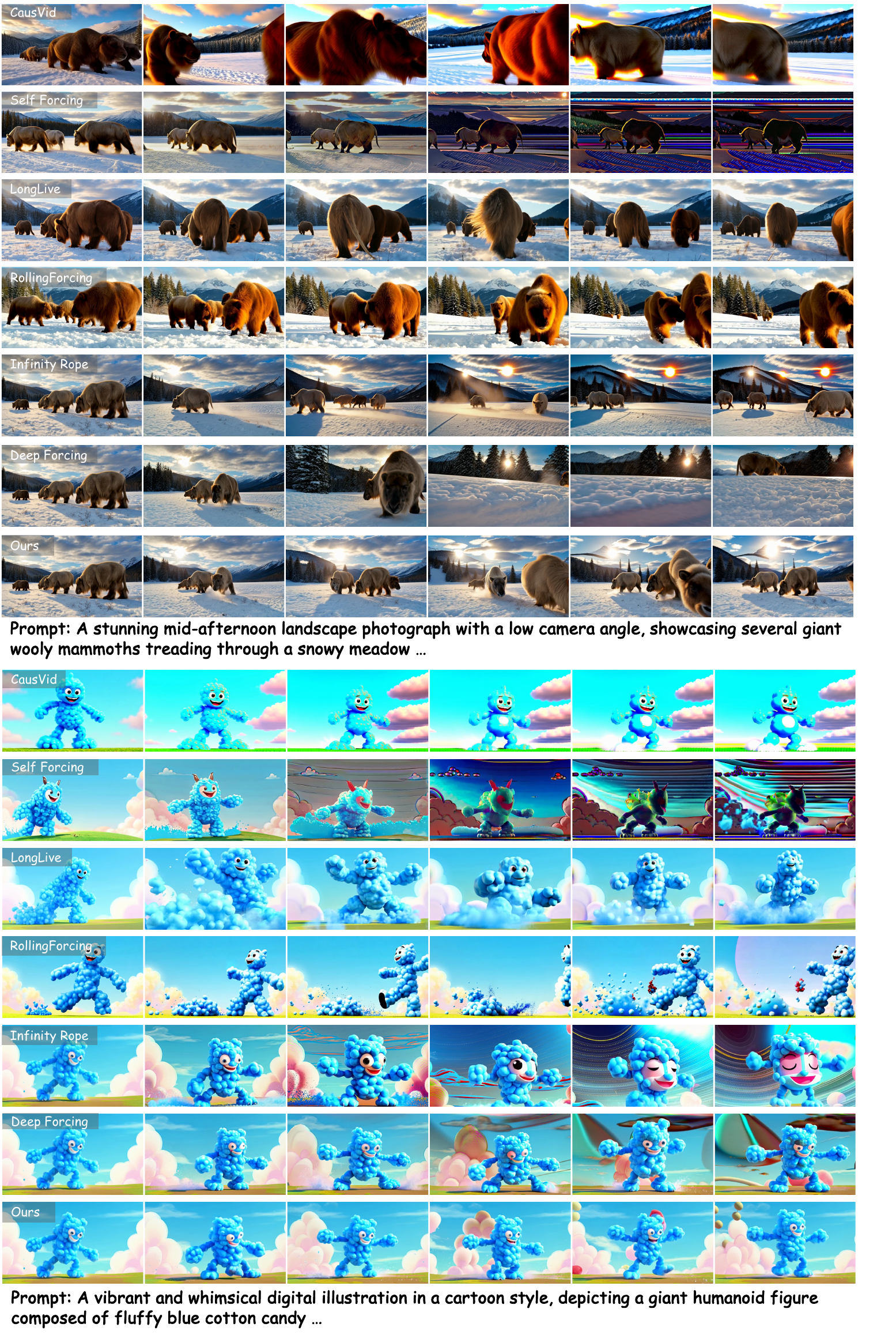}
    \caption{\textbf{Qualitative results on single-prompt long video generation (60\,s)}}
    \label{fig:quality_2}
    \vspace{-15pt}
\end{figure}

In this section, we provide more visual results, which are organized as follows:

\noindent Fig~\ref{fig:quality_1} and Fig.~\ref{fig:quality_2} presents more qualitative results on single-prompt long video generation (60\,s).

\begin{figure}[!htbp]
    \centering
    \includegraphics[width=1.05\linewidth]{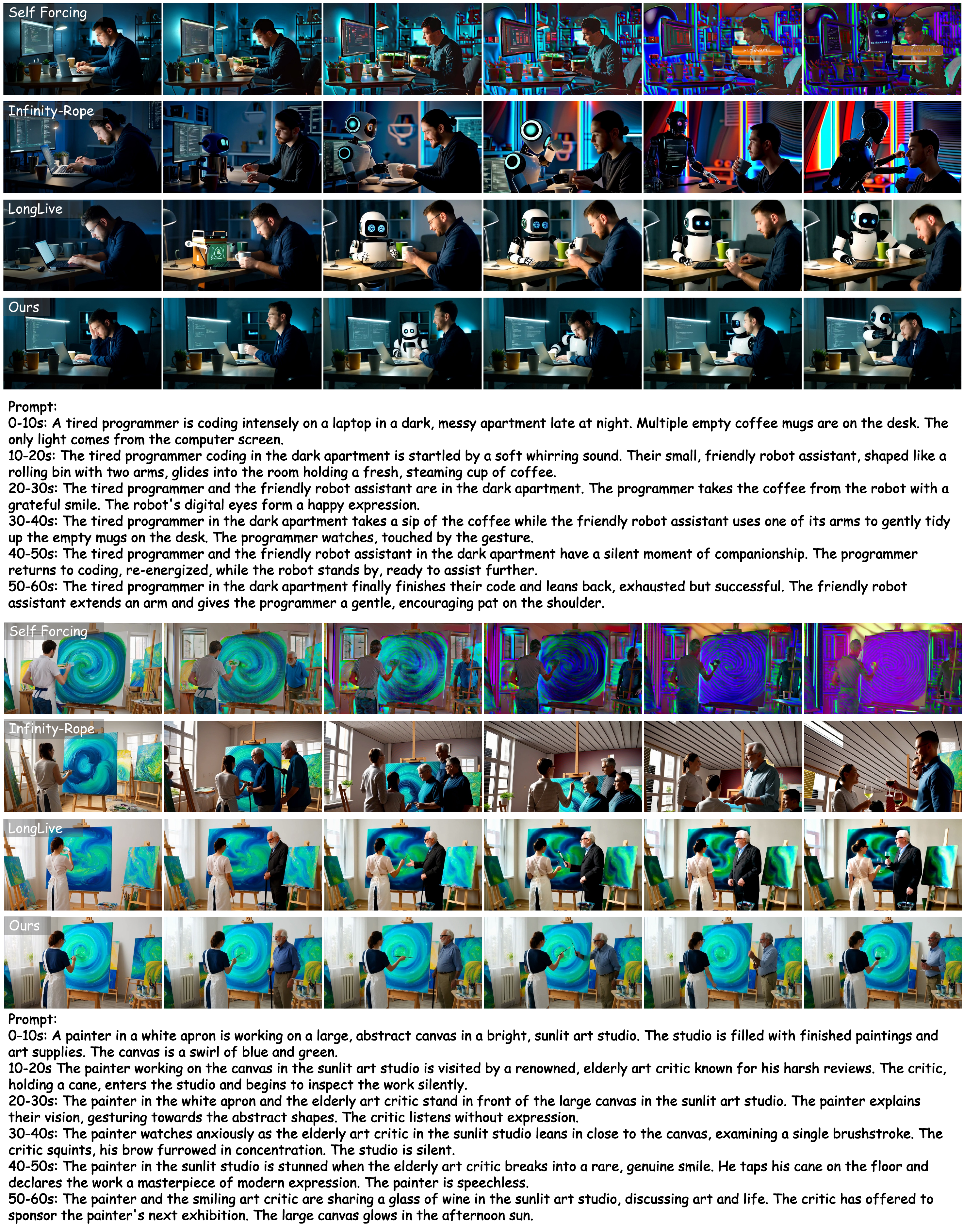}
    \caption{\textbf{Qualitative results on multi-prompt long video generation (60\,s)}}
    \label{fig:multi_1}
    \vspace{-15pt}
\end{figure}

\begin{figure}[!htbp]
    \centering
    \includegraphics[width=1.05\linewidth]{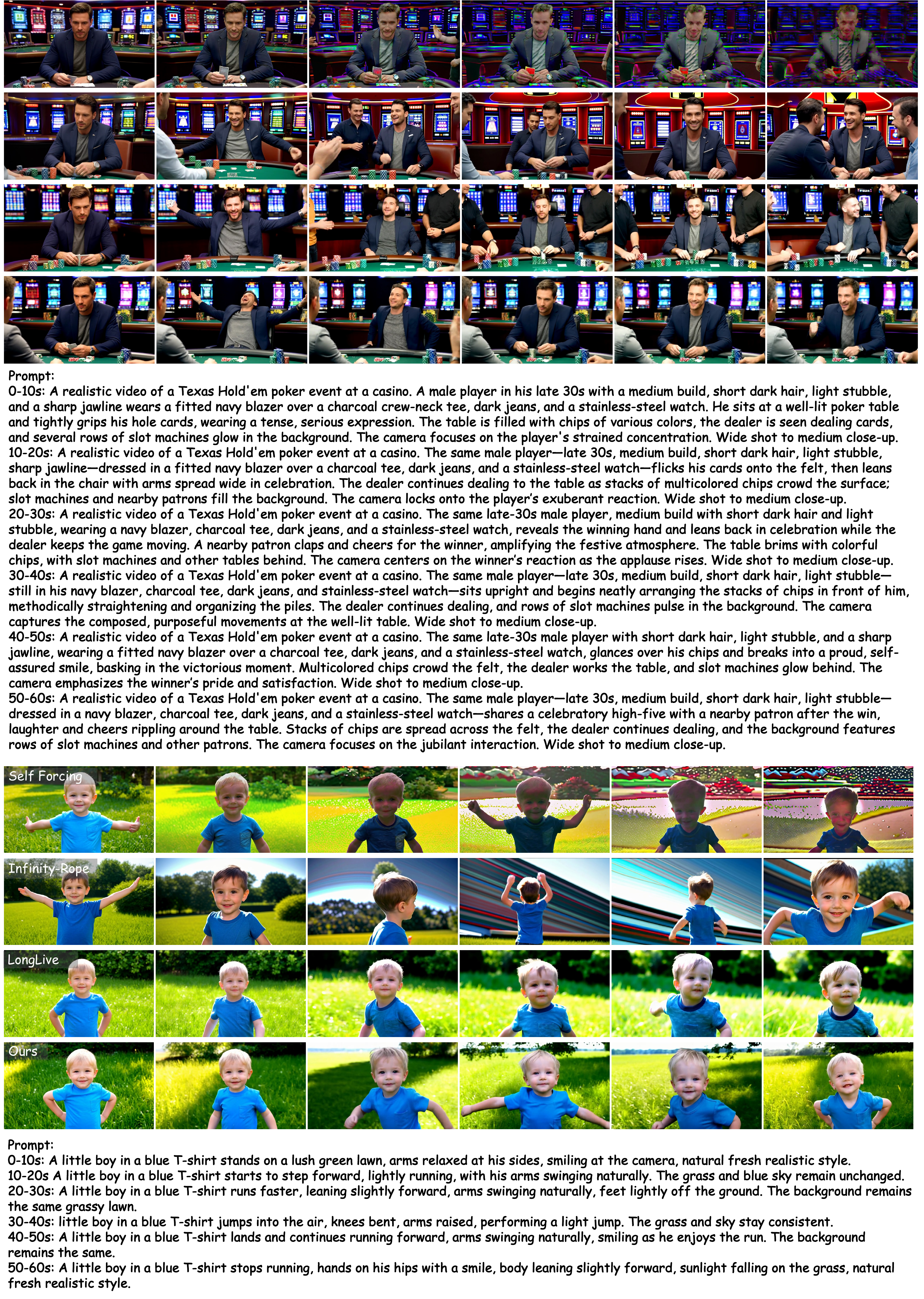}
    \caption{\textbf{Qualitative results on multi-prompt long video generation (60\,s)}}
    \label{fig:multi_2}
    \vspace{-15pt}
\end{figure}

\noindent Fig~\ref{fig:multi_1} and Fig.~\ref{fig:multi_2} presents more qualitative results on multi-prompt long video generation (60\,s).

\begin{figure}[!htbp]
    \centering
    \includegraphics[width=1.05\linewidth]{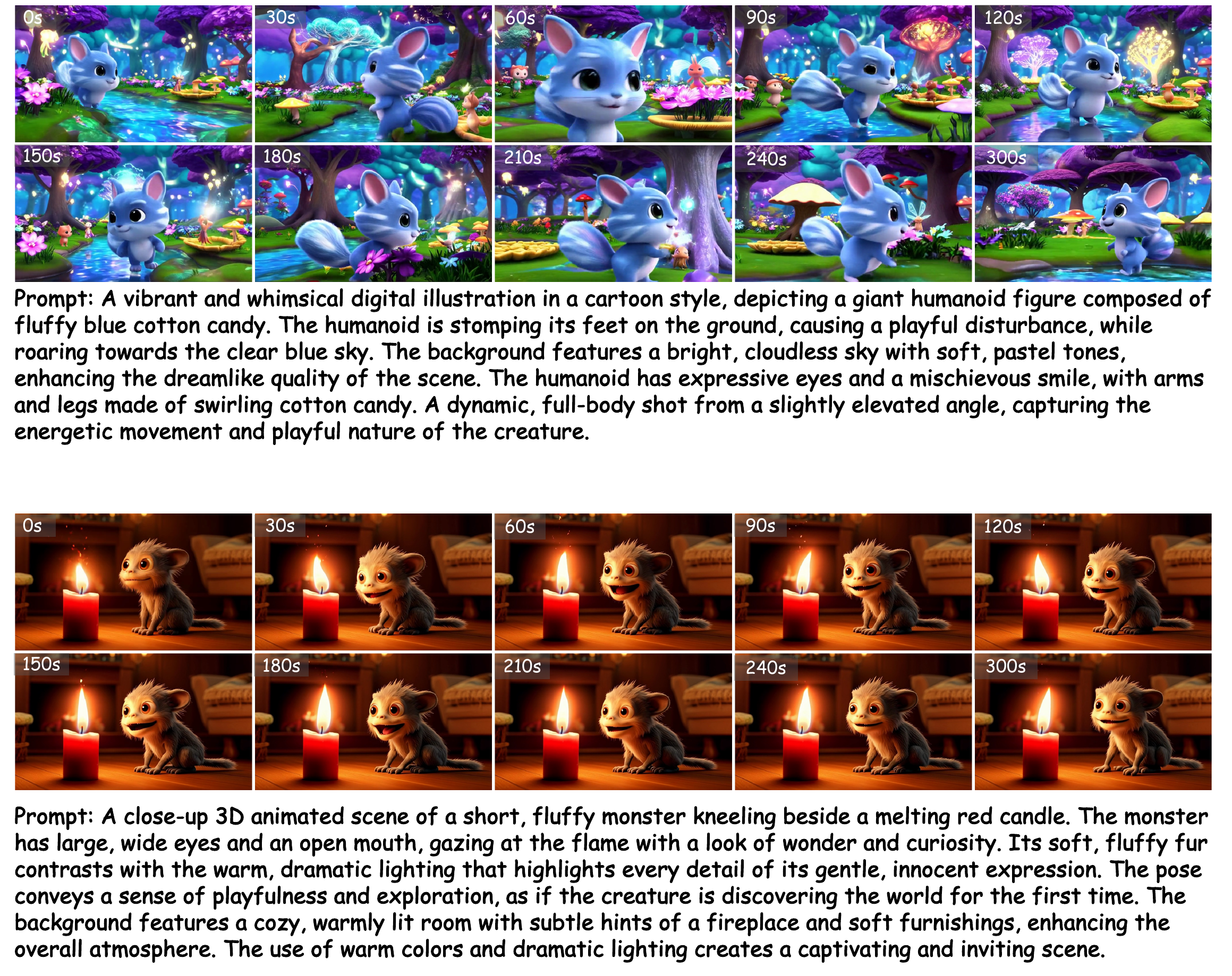}
    \caption{\textbf{Visual results on ultra long video generation (5min)}}
    \label{fig:ultra_long}
    \vspace{-15pt}
\end{figure}

\noindent Fig~\ref{fig:ultra_long} presents more visual results on ultra long video generation (5min).

